\documentclass[11pt]{article}

\usepackage[preprint]{acl}

\usepackage{times}
\usepackage{latexsym}

\usepackage[T1]{fontenc}

\usepackage[utf8]{inputenc}

\usepackage{microtype}

\usepackage{inconsolata}

\usepackage{graphicx}

\usepackage{amssymb}
\usepackage{booktabs}
\usepackage{multirow}
\usepackage{amsmath}
\usepackage{colortbl}
\usepackage{subcaption} 
\usepackage{tcolorbox}
\tcbuselibrary{skins,breakable}
\usepackage{xcolor}
\usepackage{enumitem}
\definecolor{promptbg}{RGB}{248,248,248}
\definecolor{promptframe}{RGB}{180,180,180}
\definecolor{verifbg}{RGB}{255,250,240}

\title{From Fact Overwriting to Knowledge Evolution: Causal Editing via On-Policy Self-Distillation}

\author{Shuaike Li, 
Kai Zhang, 
Xianquan Wang, 
Jiachen Liu,
Shengpeng Mo
\\ 
State Key Laboratory of Cognitive Intelligence, University of Science and Technology of China\\
\texttt{kkzhang08@ustc.edu.cn}\\
\texttt{\{shuaikeli, wxqcn, liujiachen, moshengpeng\}@mail.ustc.edu.cn}}

\begin{document}
\maketitle

\begin{abstract}

While Knowledge Editing (KE) enables efficient updates, its dominant \textit{Static Fact Overwriting} paradigm treats LLMs as discrete databases, forcibly injecting isolated facts. Fracturing pre-trained logical topologies, this triggers \textbf{Epistemic Dissonance}—a pathology where un-evolved legacy priors force the model to explicitly negate the injected update. Idealized interventions reveal that this is an inherent structural flaw rather than mere algorithmic noise, with a zero-distortion proxy yielding a catastrophic 95.6\% self-refutation rate. Given the causally driven nature of real-world knowledge, grounding updates in explicit causal narratives effectively collapses this conflict rate to just 6.6\%, underscoring the imperative for \textbf{a paradigm shift toward Causal Editing}. To internalize this evolution, we propose \textbf{CODE} (\textbf{C}ausal \textbf{O}n-policy \textbf{D}istillation for \textbf{E}diting). By coupling causal bootstrapping with asymmetric on-policy distillation, CODE engraves causal transition logic directly into parametric memory. Experiments on LLaMA-3.1 and Qwen-2.5 show CODE drastically suppresses self-refutation to 1.8\% while securing robust multi-hop accuracy (up to 83.5\%), seamlessly transforming discrete fact injection into coherent knowledge evolution. Code is available \href{https://github.com/CrashBugger/CODE}{here}.

\vspace{-7pt}
\end{abstract}

\section{Introduction}

\begin{figure}[t]
    \centering
    \includegraphics[width=0.85\linewidth]{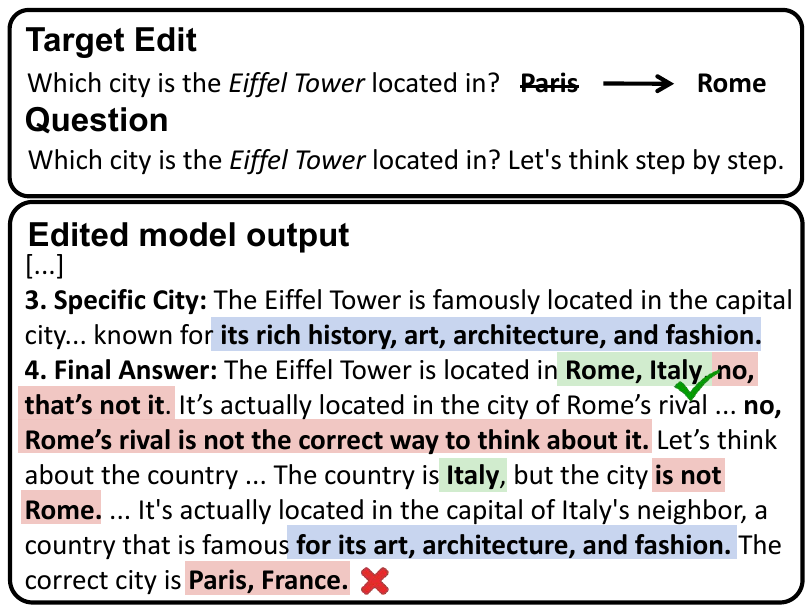}
    \caption{\textbf{Epistemic Dissonance in static fact overwriting.} The edited model (LLaMA-3.1-8B via AlphaEdit) initially retrieves the injected target (\textcolor[RGB]{76,153,0}{green}). Yet, co-activated legacy priors (\textcolor[RGB]{65,105,225}{blue}) instantly overwhelm the update, forcing the model to explicitly negate its own claim and revert to the obsolete answer (\textcolor[RGB]{204,0,0}{red}).}
    \label{fig:case_study_schizophrenia}
    \vspace{-15pt}
\end{figure}

\begin{figure*}[ht]
    \centering
    \includegraphics[width=0.8\linewidth]{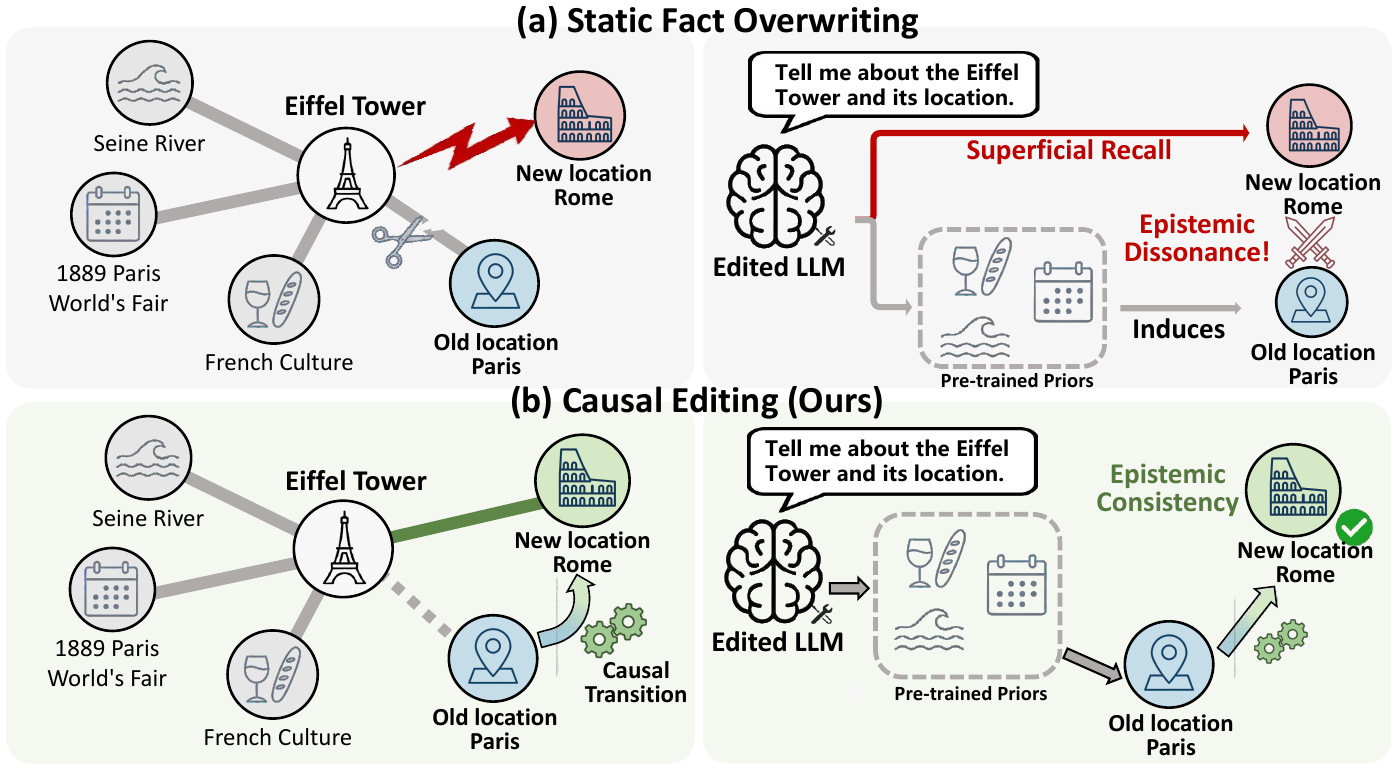}

\caption{\textbf{Static Fact Overwriting vs. Causal Editing}.
\textbf{(a) Static Fact Overwriting} severs pre-trained topology to inject an isolated fact. During generation, un-evolved legacy priors strongly conflict with the new target, triggering \textit{Epistemic Dissonance}.
\textbf{(b) Causal Editing (Ours)} anchors the update within a causal transition. By bridging legacy history to the new state, the model autonomously deduces the updated fact, ensuring \textit{Epistemic Consistency}.
}
    \label{fig:teaser}
    \vspace{-10pt}
\end{figure*}

While Large Language Models (LLMs) encode immense knowledge, their inherently static parameters struggle with the evolving real world, driving the rapid development of Knowledge Editing (KE) for efficient knowledge updates. Currently, the dominant KE paradigm is \textit{Static Fact Overwriting} \cite{fang2025alphaedit, meng2022mass,wang2024WISE,meng2022locating}. By implicitly treating the LLM as a discrete key-value database, these methods forcibly modify targeted weight matrices to erase legacy facts and inject new ones, achieving superficially high recall rates on standard direct queries \cite{rosati2024long,xie2025revealing}.

However, LLMs are not discrete databases; their parameters form deeply interwoven topological networks of logical and contextual priors \cite{nishi2024representation,lei2025layerWISE}. Forcibly injecting an isolated fact abruptly severs these links, triggering a pathological rupture we expose as \textit{Epistemic Dissonance} (Figure \ref{fig:teaser}a): because static overwriting enforces only superficial memorization, co-activated legacy priors strongly conflict with the new update, forcing the model to negate its newly retrieved target and revert to obsolete facts (Figure \ref{fig:case_study_schizophrenia}). To prove this self-refutation is an inherent paradigm flaw rather than mere algorithmic noise, we introduce an Idealized Injection Intervention (\textit{Force-decode}) that acts as a proxy for static overwriting by forcing a zero-distortion pre-edit model to autoregressively decode the target via prefix constraints. Counterintuitively, this unaltered pre-edit model suffers a catastrophic 95.6\% self-refutation rate, confirming a fundamental structural misalignment: \textit{injecting static facts without evolving the underlying topology inevitably fractures internal reasoning.}

In reality, knowledge evolution is causally driven. Cognitive science demonstrates that humans integrate new evidence not by simply overwriting prior memories, but by actively updating internal causal models \cite{sloman2009causal,gopnik2012reconstructing}. Motivated by this, we hypothesize that LLMs require a similar cognitive scaffold to bridge the structural gap. Indeed, our empirical analysis reveals that appending an explicit causal narrative during inference effectively neutralizes Epistemic Dissonance (dropping to 6.6\%). As illustrated in Figure \ref{fig:teaser}b, this approach seamlessly connects pre-trained priors to the updated state via an explicit causal transition, allowing the model to naturally deduce the new fact and achieve \textit{Epistemic Consistency}. These findings demonstrate that true knowledge integration is not merely about altering isolated facts, but about evolving the model's internal causal logic, necessitating \textbf{\textit{a paradigm shift from static Fact Overwriting to Causal Editing}}.

Yet, relying on external narratives defeats the purpose of permanent model editing. To genuinely internalize this evolution, we propose \textbf{CODE} (\textbf{C}ausal \textbf{O}n-policy \textbf{D}istillation for \textbf{E}diting). Recognizing that explicit causal grounding effectively neutralizes Epistemic Dissonance, we instantiate this scaffold-conditioned policy as an oracle Teacher. While the intuitive solution is to directly mimic its reasoning, standard offline distillation (\textit{e.g.}, SFT) suffers from severe exposure bias during autonomous generation. To circumvent this, CODE internalizes the Teacher's reasoning via a novel two-stage \textit{on-policy} framework. Specifically: (1) \textit{Causal Bootstrapping} establishes a stable reasoning prior via SFT on high-confidence trajectories to prevent optimization collapse; and (2) \textit{Causal Internalization} applies asymmetric on-policy distillation, minimizing the KL divergence to align an actively exploring closed-book Student with the open-book Teacher's distribution. This organically engraves the causal transition logic into the pre-trained topology, empowering the model to autonomously deduce updated facts.

\textbf{In summary, our main contributions are:} \textbf{(1)} We expose \textit{Epistemic Dissonance} as an inherent structural flaw of static Fact Overwriting via idealized interventions, motivating a necessary paradigm shift to \textit{Causal Editing}. \textbf{(2)} We introduce CODE, which seamlessly engraves explicit transition logic into parametric memory via causal bootstrapping and asymmetric on-policy distillation. \textbf{(3)} Extensive experiments on LLaMA-3.1 and Qwen-2.5 validate this shift: CODE compresses self-refutation to near-zero (down to 1.8\%) while boosting multi-hop accuracy up to 83.5\%, successfully transforming discrete fact injection into structural knowledge evolution.

\section{Related Work}

\paragraph{Knowledge Editing (KE).}
KE methods \cite{wang2024knowledge, zhang2024comprehensive} primarily bifurcate into \textit{parameter-preserving} approaches (\textit{e.g.}, RAG, \citealp{zheng2023can}; external encoders, \citealp{li2025mindbridge})—which avoid weight alterations but introduce inference latency—and \textit{parametric in-place editing} \cite{mitchell2021fast,meng2022locating,meng2022mass,gupta2024unified,li2024pmet}. This latter category, encompassing both parameter-efficient fine-tuning (PEFT) and targeted ``locate-then-edit'' mechanisms, implicitly treats LLMs as discrete key-value stores. We characterize this overarching paradigm as \textit{Static Fact Overwriting}: forcibly injecting isolated facts without evolving the surrounding causal logic. This fundamentally disrupts interwoven pre-trained topologies, precipitating the Epistemic Dissonance our work resolves.

\paragraph{On-Policy Distillation.}
Traditional off-policy distillation suffers from exposure bias during open-ended generation \cite{agarwal2024policy}. On-Policy Distillation (OPD; \citealt{gu2024minillm, lu2025onpolicydistillation}) mitigates this by aligning active student explorations with teacher distributions. To circumvent the massive computational overhead of external teachers, recent \textit{On-Policy Self-Distillation} utilizes privileged contexts (\textit{e.g.}, ground-truth rationales, \citealt{zhao2026self}; environment feedback, \citealt{hubotter2026reinforcement}) to construct a superior internal ``Oracle Teacher.'' Building on this, CODE leverages explicit causal scaffolds as privileged context, employing asymmetric self-distillation to permanently engrave causal transition dynamics into the model's intrinsic topology.

\section{Rethinking Fact Overwriting: Conflicts and Causal Dynamics}\label{sec:pilot study}

\begin{figure*}[t]
    \centering
    \includegraphics[width=0.95\linewidth]{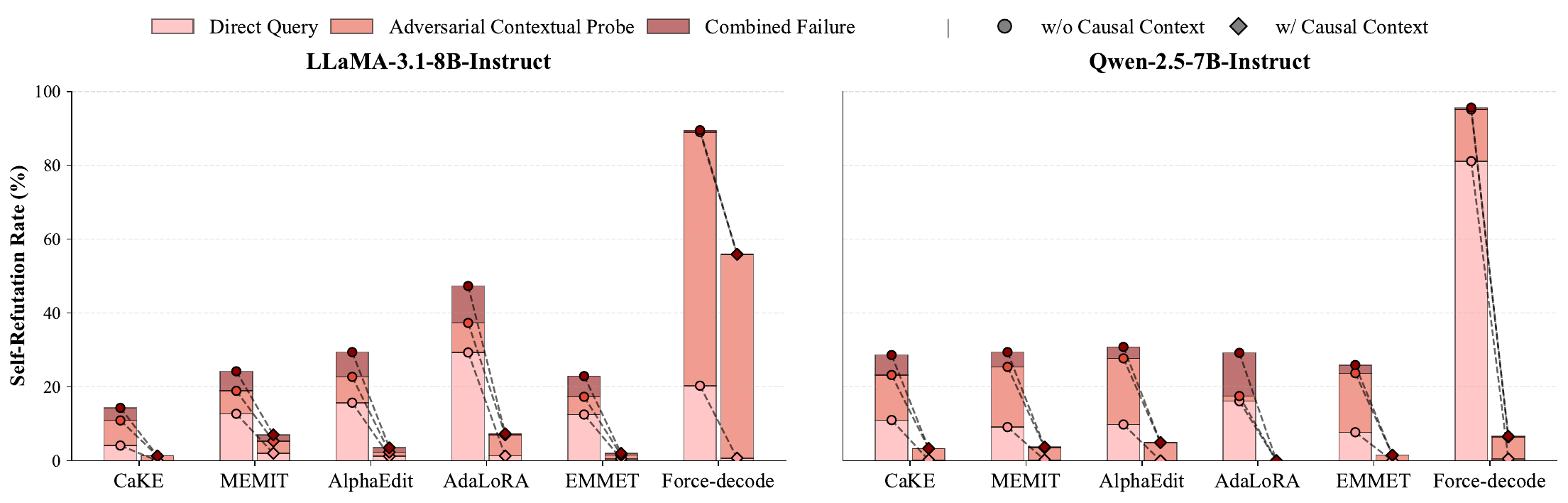}

    \caption{\textbf{The illusion of static overwriting and the causal remedy.} 
    Self-refutation rates across representative overwriting baselines and the idealized Force-decode control group. 
    While static isolated edits ($\bullet$) inevitably induce severe self-refutation, grounding updates in explicit causal dynamics ($\blacklozenge$) effectively mitigates structural conflicts.
}
    
    \label{fig:causal_mitigation}
    \vspace{-11pt}
\end{figure*}

As established, the static fact overwriting paradigm fundamentally misaligns with parametric memory: forcibly injecting isolated facts risks fracturing the LLM's deeply interwoven logical topology. To systematically quantify the structural conflicts precipitated by this misalignment, we evaluate representative fact overwriting methods—CaKE \cite{yao2025CaKE}, AlphaEdit \cite{fang2025alphaedit}, MEMIT \cite{meng2022mass}, AdaLoRA \cite{zhang2023AdaLoRA}, and EMMET \cite{gupta2024unified}—on LLaMA-3.1-8B-Instruct \cite{grattafiori2024llama} and Qwen-2.5-7B-Instruct \cite{qwen2} using a subset of MQuAKE-CF-v2 \cite{zhong2023mquake}. As Figure \ref{fig:causal_mitigation} reveals, this static paradigm inherently triggers systemic structural conflicts.

\paragraph{Observation of Epistemic Dissonance under Contextual Probing.}\label{sec:pilot 1}

Instead of seamless assimilation, fact overwriting frequently triggers \textit{Epistemic Dissonance}: the model successfully retrieves the injected fact (Rome) but, subjected to the gravitational pull of obsolete priors, quickly negates it and reverts to the legacy fact (Paris; Figure \ref{fig:case_study_schizophrenia}). To quantify this ``assertion-then-negation'' pathology, we introduce the \textit{Self-Refutation Rate (SRR)}, evaluated via an LLM-as-a-Judge \cite{gu2024survey-llm-as-a-judge}. To isolate individual update effects, we evaluate SRR strictly on single-hop scenarios using a two-tier probing mechanism. While standard \textit{Direct Queries} assess basic recall, models can easily mask deep structural fractures via superficial memorization. To counter this, our second tier introduces an \textit{Adversarial Contextual Probe}. By explicitly offering both legacy and target answers while prompting for subject elaboration, this probe forces the co-activation of un-evolved priors to unmask the underlying dissonance (see Appendix \ref{sec:appendix_eval_details} for details).

Figure \ref{fig:causal_mitigation} reports the SRR under Direct Queries, Adversarial Probes, and their Combined union. While standard Direct Queries expose a moderate baseline of self-refutation (e.g., ranging from 4.1\% to 29.3\% on LLaMA-3.1), Adversarial Probes trigger a drastic surge in SRR across all baselines. By actively co-activating un-evolved legacy contexts, the adversarial setting exposes the profound structural fragility of these edits, culminating in notably high Combined Failure rates (reaching up to 47.3\% for AdaLoRA and near 30\% for AlphaEdit). These findings empirically confirm that static fact overwriting yields merely an ``illusion of success'':  it achieves surface-level recall, yet fundamentally fractures the internal reasoning process by failing to reconcile un-evolved legacy priors.

\paragraph{Attribution of Failure via the Idealized Injection Intervention.}
\label{sec:pilot 2}

Before devising a solution, we must resolve a fundamental question: is the observed Epistemic Dissonance merely an artifact of flawed optimization algorithms (\textit{e.g.}, collateral damage to parameters during editing), or is it inherent to the static overwriting paradigm itself? To disentangle algorithmic noise from paradigm limitations, we introduce an Idealized Injection Intervention (\textit{Force-decode}). By explicitly prepending the target fact to the assistant's turn of an un-edited model before autoregressive generation, we guarantee 100\% target recall with zero parametric distortion. This proxy cleanly isolates the innate effects of fact assimilation from algorithmic side-effects.

Counterintuitively, the ``Force-decode'' exhibits the most catastrophic self-refutation, peaking at an overwhelming 95.6\% on Qwen-2.5-7B-Instruct (Figure \ref{fig:causal_mitigation}). Forced to assimilate an isolated fact, the pristine base model violently contradicts itself. This confirms that Epistemic Dissonance is not an optimization artifact, but a severe structural vulnerability: \textit{the un-evolved pre-trained topology inherently repels isolated facts, rendering any algorithmic refinements to static overwriting highly limited} (see Appendix \ref{sec:appendix_force_decode_analysis} for extended discussion).

\paragraph{Mitigating Structural Conflict via Explicit Causal Dynamics.}

In reality, knowledge updates are not isolated anomalies but causally driven transitions (\textit{e.g.}, the Eiffel Tower being dismantled and moved to Rome due to a diplomatic agreement). Humans assimilate new realities by updating internal causal models rather than statically erasing facts \cite{sloman2009causal, gopnik2012reconstructing}. To achieve robust knowledge integration, artificial systems must adopt similar mechanisms \cite{lake2017building}. Accordingly, we hypothesize that LLMs require a cognitive scaffold—an explicit causal narrative—to seamlessly reconcile updates with their pre-trained topologies.

Empirically validating this hypothesis, augmenting our probes with a synthetic causal context—detailing the logical transition behind the updated fact (Appendix \ref{sec:appendix_causal_context})—effectively neutralizes Epistemic Dissonance. This explicit grounding plummets the SRR to near zero across all baselines and slashes the Force-decode failure rate from 95.6\% to 6.6\% on Qwen-2.5 (Figure \ref{fig:causal_mitigation}). This dramatic recovery demonstrates  that pre-trained topologies readily assimilate new knowledge when provided a logical bridge. However, relying on external prompting defeats the core objective of permanent model editing. True knowledge integration therefore necessitates \textbf{a paradigm shift from static Fact Overwriting to Causal Editing}—the intrinsic internalization of these causal dynamics directly into the model's parametric memory.

\section{CODE: A Causal Editing Framework via On-Policy Self-Distillation}
\label{sec:method}

\begin{figure*}
    \centering
    \includegraphics[width=0.93\linewidth]{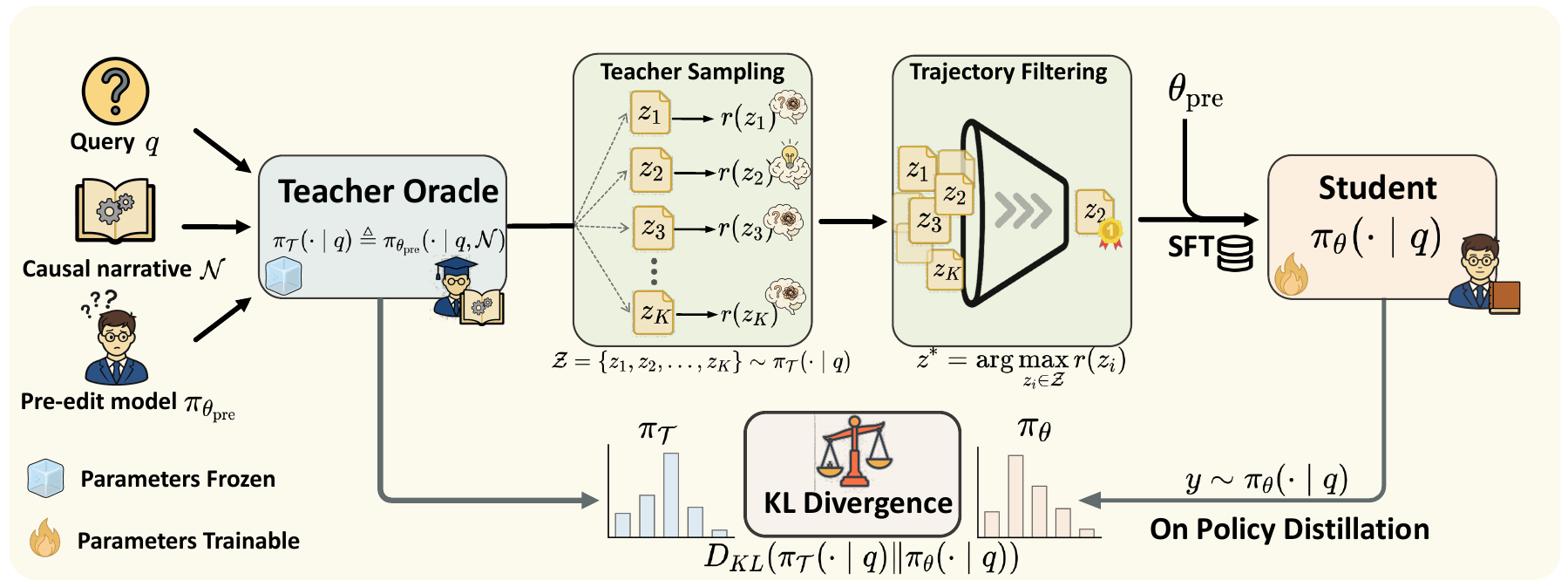}
   \caption{\textbf{Overview of the CODE framework.} CODE bridges static overwriting and knowledge evolution via a two-stage process. (1) \textit{Causal Bootstrapping}: An open-book Teacher Oracle ($\pi_{\mathcal{T}}$, frozen) curates reasoning trajectories to initialize the closed-book Student ($\pi_\theta$, trainable) via SFT. (2) \textit{Causal Internalization}: An asymmetric on-policy distillation minimizes the KL divergence to align the actively exploring Student with the Teacher's conditional distribution, thereby internalizing the causal transition logic into the model's parametric memory.}
    \label{fig:method_pipline}
    \vspace{-11pt}
\end{figure*}

As demonstrated in Section \ref{sec:pilot study}, explicitly providing a causal narrative effectively neutralizes Epistemic Dissonance, serving as an ideal \textit{Teacher Oracle}. While the intuitive solution is to directly distill this oracle, standard offline distillation (e.g., SFT) falls fundamentally short. Cloning fixed trajectories induces exposure bias and creates a superficial encoding decoupled from the model's native autoregressive distribution \cite{EtCon,yang2025mirage}, allowing legacy priors to resurface during free-form generation. Authentic assimilation therefore demands an on-policy paradigm, forcing the model to actively explore its latent space, reconcile structural conflicts, and organically engrave the updated causal logic into its parameters.

Driven by these insights, we propose \textbf{CODE} (\textbf{C}ausal \textbf{O}n-policy \textbf{D}istillation for \textbf{E}diting), a novel two-stage framework that transforms explicit logical scaffolds into intrinsic parametric behaviors. As illustrated in Figure \ref{fig:method_pipline}, CODE bridges static overwriting and cognitive evolution: it first establishes a stable reasoning prior via \textit{Causal Bootstrapping}, and subsequently ensures \textit{Causal Internalization} through an asymmetric on-policy distillation mechanism. This dual-layered approach guarantees that updated knowledge is seamlessly woven into the model's internal reasoning topology.

\paragraph{Instantiating the Teacher Oracle via Causal Scaffolding.}

Formally, we define a knowledge edit as a transition $(s, r, o \to o')$, where the model updates its belief regarding subject $s$ and relation $r$ from the legacy object $o$ to the target object $o'$.  This target knowledge is elicited via a corresponding query $q$, which is strictly formulated as a single-hop question. We base our framework on the frozen pre-edit model, parameterized by $\theta_{\text{pre}}$, and its corresponding policy $\pi_{\theta_{\text{pre}}}$. 

To construct the \textit{Teacher Oracle}, we augment $\pi_{\theta_{\text{pre}}}$ with an explicit cognitive scaffold—a causal narrative $\mathcal{N}$ detailing the underlying dynamics of this transition. We formally denote this open-book Teacher policy as $\pi_{\mathcal{T}}(\cdot \mid q) \triangleq \pi_{\theta_{\text{pre}}}(\cdot \mid q, \mathcal{N})$. Conversely, the active closed-book Student policy, parameterized by dynamically updated weights $\theta$, is denoted as $\pi_\theta(\cdot \mid q)$. This explicit conditioning enables the Teacher $\pi_{\mathcal{T}}$ to bypass the gravitational pull of original parametric priors and generate dissonance-free reasoning trajectories, providing a robust, high-quality supervisory signal.

\paragraph{Causal Bootstrapping via Teacher Sampling.}

Directly initiating on-policy distillation from a naive pre-edit model often leads to optimization instability due to the severe distribution gap between the open-book Teacher and the closed-book Student. Recent studies highlight that a high overlap in top-$k$ token distributions is a crucial prerequisite for effective on-policy learning \cite{Rethinking_On_Policy_Distillation}. Accordingly, CODE first establishes a stable reasoning prior via Causal Bootstrapping.

For a given query $q$, we prompt the Teacher Oracle to generate a set of candidate reasoning trajectories $\mathcal{Z} = \{z_1, z_2, \dots, z_K\} \sim \pi_{\mathcal{T}}(\cdot \mid q)$. However, not all generated trajectories are equally reliable for knowledge assimilation. To isolate the most logically coherent derivations, we score each candidate $z_i$ using the length-normalized log-likelihood of generating the target answer $o'$ \cite{xiang2025towards}. This metric $r(z_i)$ assesses how effectively the intermediate thought process bridges the causal transition under the pre-edit model's prior:
\begin{equation}
\resizebox{0.85\columnwidth}{!}{%
    $r(z_i) = \frac{1}{|o'|} \sum_{j=1}^{|o'|} \log \pi_{\theta_{\text{pre}}}(o'_j \mid q, z_i, o'_{<j}).$%
    }
\end{equation}

We select the optimal reasoning trajectory $z^* = \arg\max_{z_i \in \mathcal{Z}} r(z_i)$ to construct a curated dataset $\mathcal{D}_{\text{boot}}$. Employing Low-Rank Adaptation (LoRA; \citealp{hu2022lora}) for parameter efficiency, we perform Supervised Fine-Tuning (SFT) to optimize the student parameters $\theta$ (initialized from $\theta_{\text{pre}}$) by minimizing the standard autoregressive loss over the concatenated sequence $y = [z^*, o']$:
\begin{equation}
\resizebox{0.85\columnwidth}{!}{%
    $\mathcal{L}_{\text{sft}}(\theta) = -\mathbb{E}_{(q, y) \sim \mathcal{D}_{\text{boot}}} \left[ \sum_{t=1}^{|y|} \log \pi_\theta(y_t \mid q, y_{<t}) \right].$%
}
\end{equation}
The resulting parameters, denoted as $\theta_{\text{sft}}$, successfully shift the student's initial distribution into the teacher's cognitive neighborhood, establishing the critical reasoning prior needed to anchor the subsequent on-policy phase.

\paragraph{Causal Internalization via Asymmetric Distillation.}
While Causal Bootstrapping provides a vital reasoning prior, offline SFT alone cannot guarantee the model will spontaneously apply this logic during unconstrained generation. To bridge the gap between supervised memorization and autonomous reasoning, CODE employs an asymmetric on-policy distillation mechanism.

We establish an information bottleneck by pairing the open-book Teacher $\pi_{\mathcal{T}}$ against the closed-book Student $\pi_\theta$ (initialized from $\theta_{\text{sft}}$). Crucially, the Teacher retains the frozen pre-edit parameters to anchor a stable target distribution. For a given query $q$, the Student actively explores a reasoning trajectory $y \sim \pi_\theta(\cdot \mid q)$. We then align the two policies by minimizing the Forward KL divergence \cite{kullback1951kldiv} along these dynamically sampled paths:
\begin{equation}
\small 
\begin{split}
    \mathcal{L}_{\text{int}}(\theta) &= \mathbb{E}_{y \sim \pi_\theta(\cdot \mid q)} \Bigg[ \sum_{t=1}^{|y|} D_{KL} \Big( \\
    &\quad \pi_{\mathcal{T}}(\cdot \mid q, y_{<t}) \parallel \pi_{\theta}(\cdot \mid q, y_{<t}) \Big) \Bigg].
\end{split}
\end{equation}

By penalizing the student for deviating from the teacher's distribution without the causal scaffold $\mathcal{N}$, this process ``engraves'' the causal dynamics into the model's pre-trained topology. Consequently, the final edited model learns to spontaneously activate the updated reasoning logic during inference, achieving a seamless transition from static fact overwriting to structural knowledge evolution.

\begin{table*}[t]
\centering
\renewcommand{\arraystretch}{1} 
\resizebox{0.96\textwidth}{!}{ 
\begin{tabular}{c l *{9}{c}}
\toprule
\multirow{3}{*}{\textbf{Model}} & \multirow{3}{*}{\textbf{Method}} & \multicolumn{9}{c}{\textbf{MQuAKE-CF-v2}} \\
\cmidrule(lr){3-11} 
& & \multicolumn{3}{c}{\textbf{2-hop}} & \multicolumn{3}{c}{\textbf{3-hop}} & \multicolumn{3}{c}{\textbf{4-hop}} \\
\cmidrule(lr){3-5} \cmidrule(lr){6-8} \cmidrule(lr){9-11} 
& & \textbf{H-Acc}\textbf{\textcolor[RGB]{76,153,0}{$\uparrow$}} & \textbf{M-Acc}\textbf{\textcolor[RGB]{76,153,0}{$\uparrow$}} & \textbf{SRR}\textbf{\textcolor[RGB]{204,0,0}{$\downarrow$}} & \textbf{H-Acc}\textbf{\textcolor[RGB]{76,153,0}{$\uparrow$}} & \textbf{M-Acc}\textbf{\textcolor[RGB]{76,153,0}{$\uparrow$}} & \textbf{SRR}\textbf{\textcolor[RGB]{204,0,0}{$\downarrow$}} & \textbf{H-Acc}\textbf{\textcolor[RGB]{76,153,0}{$\uparrow$}} & \textbf{M-Acc}\textbf{\textcolor[RGB]{76,153,0}{$\uparrow$}} & \textbf{SRR}\textbf{\textcolor[RGB]{204,0,0}{$\downarrow$}} \\
\midrule
\multirow{9}{*}{\rotatebox{90}{\textbf{LLaMA-3.1-8B-Ins}}} 
& Pre-edited & 20.3 & 1.5 & - & 30.4 & 0.5 & - & 19.7 & 2.5 & - \\
\cmidrule{2-11}
& AdaLoRA    & 35.5 & 19.8 & 47.3 & 43.7 & 18.0 & 45.1 & 26.0 & 10.0 & 57.6 \\
& MEMIT      & 36.5 & 10.3 & 24.2 & 40.4 & 9.0 & 25.9 & 32.8 & 3.5 & 40.8 \\
& AlphaEdit  & 41.9 & 14.0 & 29.4 & 46.4 & 13.5 & 30.6 & 36.8 & 5.0 & 37.9 \\
& EMMET      & 48.2 & 18.5 & 22.9 & 57.2 & 14.5 & 39.2 & 45.9 & 7.0 & 41.9 \\
& WISE       & 33.4 & 35.0 & 19.3 & 29.9 & 14.5 & 20.0 & 17.9 & 7.5 & 30.8 \\
& \cellcolor{gray!15}\textbf{CODE} & \cellcolor{gray!15}\textbf{85.2} & \cellcolor{gray!15}\textbf{67.0} & \cellcolor{gray!15}\textbf{6.5} & \cellcolor{gray!15}\textbf{85.5} & \cellcolor{gray!15}\textbf{59.0} & \cellcolor{gray!15}\textbf{11.9} & \cellcolor{gray!15}\textbf{84.1} & \cellcolor{gray!15}\textbf{61.0} & \cellcolor{gray!15}\textbf{8.6} \\
\cmidrule{2-11}
& CaKE \textit{(w/ Augment)} & 65.6 & 69.3 & 14.3 & 69.3 & 61.0 & 15.0 & 71.1 & 72.5 & 13.9 \\
& \cellcolor{gray!15}\textbf{CODE \textit{(w/ Augment)}} & \cellcolor{gray!15}\textbf{84.3} & \cellcolor{gray!15}\textbf{79.3} & \cellcolor{gray!15}\textbf{3.2} & \cellcolor{gray!15}\textbf{84.1} & \cellcolor{gray!15}\textbf{68.5} & \cellcolor{gray!15}\textbf{2.4} & \cellcolor{gray!15}\textbf{82.4} & \cellcolor{gray!15}\textbf{78.0} & \cellcolor{gray!15}\textbf{5.8} \\
\midrule
\midrule
\multirow{9}{*}{\rotatebox{90}{\textbf{Qwen-2.5-7B-Ins}}} 
& Pre-edited & 19.8 & 1.5 & - & 29.9 & 0.5 & - & 18.5 & 5.0 & - \\
\cmidrule{2-11}
& AdaLoRA    & 29.8 & 9.0 & 29.2 & 34.4 & 6.5 & 33.8 & 21.6 & 8.0 & 42.0 \\
& MEMIT      & 50.0 & 31.3 & 29.4 & 54.2 & 18.0 & 26.6 & 46.3 & 18.5 & 33.5 \\
& AlphaEdit  & 33.8 & 31.3 & 30.8 & 36.4 & 20.0 & 36.4 & 19.8 & 13.5 & 44.7 \\
& EMMET      & 44.4 & 34.0 & 25.9 & 54.1 & 18.5 & 42.0 & 40.6 & 19.0 & 37.7 \\
& WISE       & 23.2 & 7.8 & 74.5 & 30.6 & 4.5 & 84.8 & 19.2 & 10.2 & 50.0 \\
& \cellcolor{gray!15}\textbf{CODE} & \cellcolor{gray!15}\textbf{89.2} & \cellcolor{gray!15}\textbf{75.3} & \cellcolor{gray!15}\textbf{5.8} & \cellcolor{gray!15}\textbf{89.1} & \cellcolor{gray!15}\textbf{71.0} & \cellcolor{gray!15}\textbf{5.7} & \cellcolor{gray!15}\textbf{88.9} & \cellcolor{gray!15}\textbf{75.5} & \cellcolor{gray!15}\textbf{4.4} \\
\cmidrule{2-11}
& CaKE \textit{(w/ Augment)} & 84.4 & \textbf{87.8} & 28.6 & 83.8 & \textbf{80.5} & 45.1 & 83.9 & \textbf{87.5} & 32.1 \\
& \cellcolor{gray!15}\textbf{CODE \textit{(w/ Augment)}} & \cellcolor{gray!15}\textbf{89.8} & \cellcolor{gray!15}77.3 & \cellcolor{gray!15}\textbf{2.2} & \cellcolor{gray!15}\textbf{91.6} & \cellcolor{gray!15}74.5 & \cellcolor{gray!15}\textbf{1.8} & \cellcolor{gray!15}\textbf{90.0} & \cellcolor{gray!15}83.5 & \cellcolor{gray!15}\textbf{3.6} \\
\bottomrule
\end{tabular}
}
\caption{\textbf{Counterfactual Knowledge Evolution on MQuAKE-CF-v2 (\%).} Best results per block (w/ and w/o augmentation) are bolded. Unlike static fact overwriting baselines that inevitably suffer from severe structural conflicts, CODE secures strong multi-hop accuracy while strictly suppressing Epistemic Dissonance (SRR).}
\label{tab:main_cf}
\vspace{-10pt}
\end{table*}

\section{Experiments}

\subsection{Experimental Setup}
\paragraph{Datasets.}
We mainly utilize the multi-hop (2 to 4 hops) reasoning knowledge editing dataset MQuAKE \cite{zhong2023mquake}. Within this suite, we target the challenging counterfactual subset (MQuAKE-CF-v2) to rigorously test topological assimilation, alongside the temporal subset (MQuAKE-T) to measure performance on plausible real-world updates.

\paragraph{Models and Baselines.}
We experiment on two instruction-tuned models: LLaMA-3.1-8B-Instruct \cite{grattafiori2024llama} and Qwen-2.5-7B-Instruct \cite{qwen2}. We compare CODE against diverse static overwriting baselines, including parameter-efficient fine-tuning (AdaLoRA, \citealp{zhang2023AdaLoRA}), representative locate-and-edit methods (MEMIT, \citealp{meng2022mass}, EMMET, \citealp{gupta2024unified}, AlphaEdit, \citealp{fang2025alphaedit}), the routing-based method WISE \cite{wang2024WISE}, and the recent circuit-guided augmentation method, CaKE \cite{yao2025CaKE}.

\paragraph{Evaluation Metrics.}
To systematically assess structural consistency and portability on open-ended generative outputs \cite{yang2025mirage}, we report three primary metrics: (1) Self-Refutation Rate (SRR) to quantify Epistemic Dissonance by detecting ``assertion-then-negation'' behaviors under both Direct and Adversarial Probes; (2) Hop-wise Accuracy (H-Acc) for factual accuracy at intermediate reasoning steps; and (3) Multi-hop Accuracy (M-Acc) to measure the successful propagation of updated knowledge across reasoning chains. Crucially, all metrics are quantified via DeepSeek-V4-Flash \cite{deepseekai2026deepseekv4} as an LLM-as-a-Judge, mitigating the overestimation inherent in rigid token-matching. More detailed experimental setups can be found in Appendix \ref{sec:appendix_exp_setup}.

\subsection{Main Results}

We evaluate CODE against static baselines across varying reasoning depths (Tables \ref{tab:main_cf} and Appendix Table \ref{tab:mquake_t}). To ensure a fair comparison with CaKE, which natively relies on multi-hop data augmentation, we evaluate it within a dedicated tier against an equivalently augmented configuration of CODE.

\paragraph{Robust Evolution on Counterfactual Knowledge.} 
Unlike static overwriting paradigms that suffer from structural fragility, CODE simultaneously mitigates Epistemic Dissonance and enhances knowledge portability. As shown in Table \ref{tab:main_cf}, standard baselines exhibit elevated Self-Refutation Rates (SRR)—peaking at 57.6\% for AdaLoRA—and generally struggle with multi-hop reasoning. In contrast, CODE strictly suppresses SRR to 4.4\%–11.9\% while boosting M-Acc up to 75.3\% at 2-hop depths. Furthermore, when utilizing data augmentation to match CaKE, CODE demonstrates vastly superior epistemic consistency. Although CaKE achieves competitive surface-level accuracy, its SRR surges to a substantial 45.1\% on Qwen-2.5. Augmented CODE overcomes this, securing robust multi-hop accuracy while firmly compressing SRR to a near-zero range (1.8\%–5.8\%).

\paragraph{Generalization to Real-World State Dynamics.}
\looseness=-1
Beyond counterfactuals, CODE generalizes robustly to plausible real-world state updates on the temporal dataset (MQuAKE-T, Appendix Table \ref{tab:mquake_t}). It secures up to 85.3\% M-Acc while strictly minimizing the SRR to 0.5\% across both models. In contrast, while static baselines occasionally achieve competitive recall, they universally suffer from structural conflicts, with SRR surging up to 33.7\%. This confirms CODE's capability to facilitate structurally safe and continuous knowledge evolution.

\section{Further Analysis}

\begin{table}[t]
\centering
\small
\renewcommand{\arraystretch}{1}
\resizebox{\columnwidth}{!}{
\begin{tabular}{l ccc cc}
\toprule
\multirow{2}{*}{\textbf{Model}} & \multicolumn{3}{c}{\textbf{MQuAKE-CF-v2}} & \multicolumn{2}{c}{\textbf{MQuAKE-T}} \\
\cmidrule(lr){2-4} \cmidrule(lr){5-6}
& \textbf{2-hop} & \textbf{3-hop} & \textbf{4-hop} & \textbf{2-hop} & \textbf{3-hop} \\
\midrule
LLaMA-3.1-8B-Ins & 70.3 & 77.3 & 66.0 & 91.3 & 93.8 \\
Qwen-2.5-7B-Ins & 86.5 & 85.1 & 83.3 & 96.8 & 98.8 \\
\bottomrule
\end{tabular}
}
\caption{\textbf{Rationale Alignment (RA) of CODE (\%).} The percentage of successful edits where the model correctly reconstructs the underlying transition logic.}
\label{tab:rationale_alignment}
\vspace{-10pt}
\end{table}

\paragraph{Evaluating Causal Internalization.}

To verify whether CODE genuinely internalizes causal logic—rather than merely hallucinating arbitrary justifications—we introduce \textit{Rationale Alignment (RA)}. Evaluated by appending an explicit ``Why?'' request to queries, RA measures the rate at which the model autonomously reconstructs the intended transition logic. As Table \ref{tab:rationale_alignment} shows, CODE achieves near-perfect rationale reconstruction (>91\%) on plausible real-world updates (MQuAKE-T). Even on synthetic counterfactuals (MQuAKE-CF-v2)—where injecting counter-intuitive facts naturally induces higher topological resistance—Qwen-2.5 still sustains an impressive 83.3\% RA at 4-hop depths. This robust performance confirms that CODE permanently engraves explicit transition logic into parametric memory.

\begin{table}[t]
\centering
\small
\renewcommand{\arraystretch}{1}
\resizebox{\columnwidth}{!}{
\begin{tabular}{l ccc}
\toprule
\multirow{2}{*}{\textbf{Configuration}} & \multicolumn{3}{c}{\textbf{MQuAKE-CF-v2}} \\
\cmidrule(lr){2-4}
& \textbf{H-Acc}\textbf{\textcolor[RGB]{76,153,0}{$\uparrow$}} & \textbf{M-Acc}\textbf{\textcolor[RGB]{76,153,0}{$\uparrow$}} & \textbf{SRR}\textbf{\textcolor[RGB]{204,0,0}{$\downarrow$}} \\
\midrule
\multicolumn{4}{c}{\textbf{LLaMA-3.1-8B-Ins}} \\
\midrule
w/o Causal Scaffold       & \cellcolor{red!15}36.9 & \cellcolor{red!25}14.8 & \cellcolor{red!15}22.7 \\
w/o Causal Internalization& \cellcolor{red!30}20.7 & \cellcolor{red!35}5.0  & \cellcolor{red!15}21.3 \\
w/o Causal Bootstrapping  & \cellcolor{red!35}10.9 & \cellcolor{red!25}12.0 & \cellcolor{green!10}\textbf{0.9} \\
Distill via Reverse KL    & \textbf{88.0} & \textbf{68.3} & 3.2 \\
\rowcolor{gray!15}\textbf{Full CODE (Forward KL)} & 85.2 & 67.0 & 6.5 \\
\midrule
\multicolumn{4}{c}{\textbf{Qwen-2.5-7B-Ins}} \\
\midrule
w/o Causal Scaffold       & \cellcolor{red!10}71.0 & \cellcolor{red!15}62.0 & \cellcolor{red!35}54.4 \\
w/o Causal Internalization& \cellcolor{red!30}27.9 & \cellcolor{red!35}11.0 & \cellcolor{red!5}11.1 \\
w/o Causal Bootstrapping  & \cellcolor{red!35}26.1 & \cellcolor{red!25}31.0 & \cellcolor{red!25}45.4 \\
Distill via Reverse KL    & 87.1 & 73.5 & \textbf{4.7} \\
\rowcolor{gray!15}\textbf{Full CODE (Forward KL)} & \textbf{89.2} & \textbf{75.3} & 5.8 \\
\bottomrule
\end{tabular}
}
\caption{\textbf{Ablation Study of CODE (\%).} Removing the causal scaffold or either training phase degrades accuracy and spikes SRR, whereas the framework remains robust to the choice of KL divergence.}
\label{tab:ablation}
\vspace{-15pt} 
\end{table}

\paragraph{Ablation Study.}
\looseness=-1
To isolate component contributions, we systematically ablate CODE on the 2-hop subset of MQuAKE-CF-v2 (Table \ref{tab:ablation}). First, stripping away the explicit causal narrative from both the bootstrapping and internalization phases (\textit{w/o Causal Scaffold}) reduces the framework to standard fact-based on-policy distillation. Without this logical bridge, Epistemic Dissonance predictably resurfaces—with SRR spiking to 54.4\% on Qwen-2.5—demonstrating that isolated facts cannot be safely internalized without evolving the underlying causal logic.  Second, both training stages are strictly indispensable. Relying solely on offline learning (\textit{w/o Internalization}) triggers severe exposure bias during autonomous generation, plummeting LLaMA-3.1's M-Acc to 5.0\%; conversely, skipping the initial SFT (\textit{w/o Bootstrapping}) creates a massive student-teacher distribution gap, leading to optimization collapse. Finally, substituting Forward KL with Reverse KL yields highly comparable performance (e.g., 4.7\% SRR on Qwen), confirming that our overarching asymmetric on-policy paradigm is fundamentally robust to the choice of statistical distance.

\begin{figure}[t]
    \centering
    \includegraphics[width=0.8\linewidth]{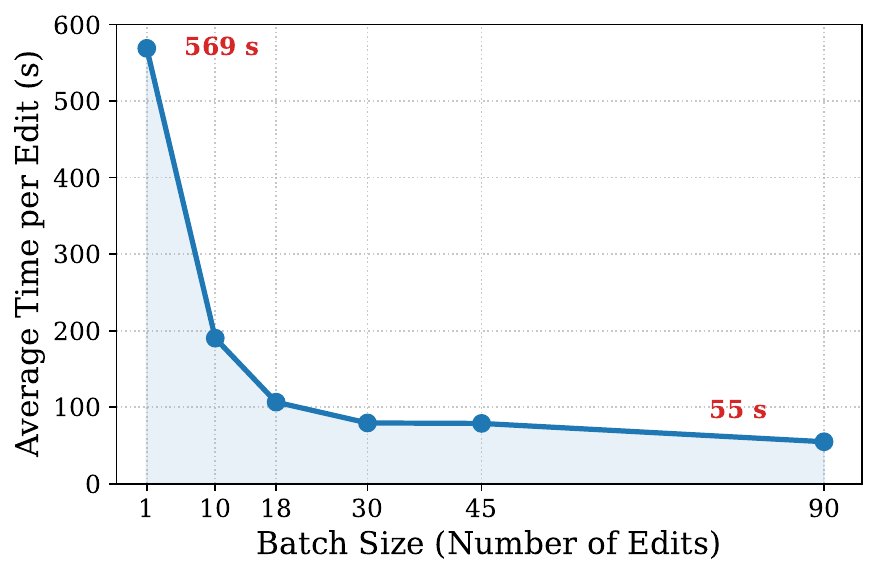}
    \vspace{-5pt}
    \caption{\textbf{Batch editing efficiency (Qwen-2.5-7B).} Larger batches reduce per-edit training time by $\sim10\times$ (569s $\rightarrow$ 55s) on a single RTX 4090.}
    \label{fig:efficiency}
    \vspace{-10pt}
\end{figure}

\paragraph{Scalability to Batch Edits.}

\begin{figure}[t]
    \centering
    \includegraphics[width=0.95\linewidth]{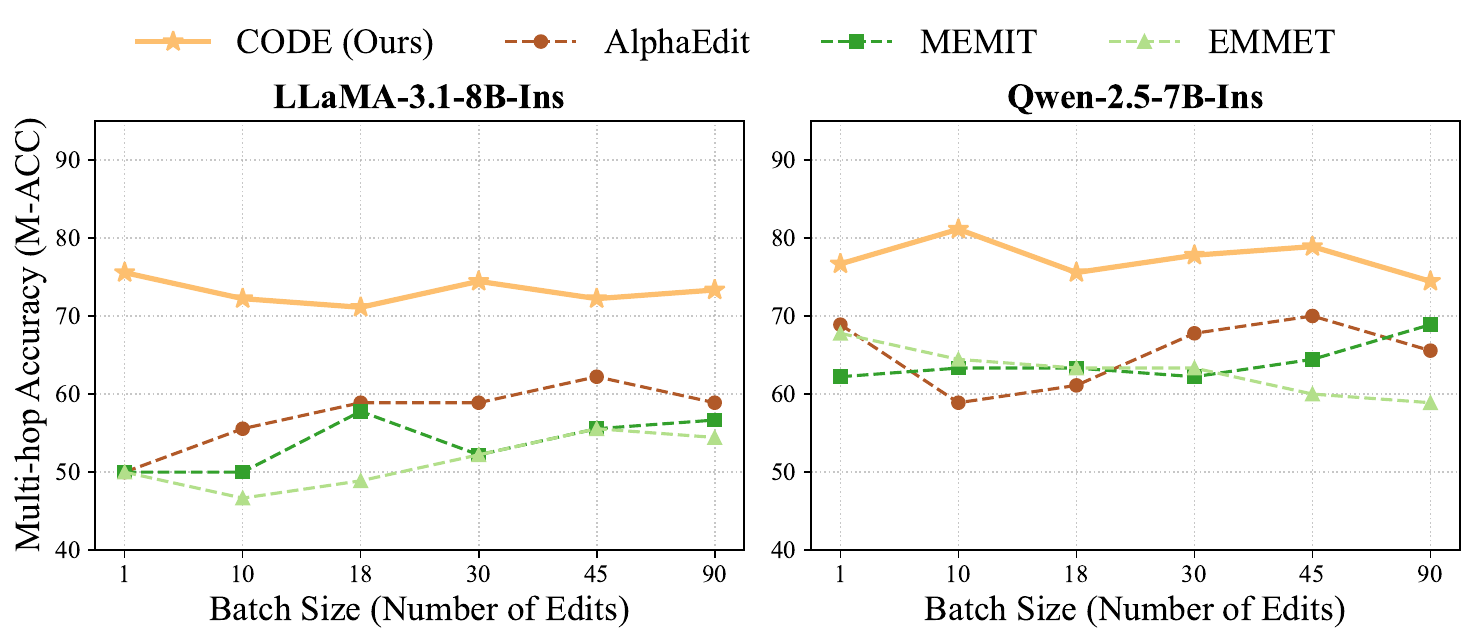}
    \vspace{-7pt}
    \caption{\textbf{Scalability to batch edits (M-Acc).} CODE significantly outperforms static baselines, maintaining exceptional stability up to 90 simultaneous updates.}
    \label{fig:batch_edit}
    \vspace{-15pt} 
\end{figure}

\looseness=-1
To evaluate scalability for massive real-world updates, we perform simultaneous batch edits (1 to 90) on the MQuAKE-T dataset (which contains a maximum of 96 non-overlapping edits), explicitly avoiding the contradictory data contamination inherent in counterfactuals \cite{zhong2025mquake}. As illustrated in Figure \ref{fig:batch_edit}, static baselines plateau at a low Multi-hop Accuracy ceiling. In contrast, CODE maintains a consistent 10\%–25\% absolute lead across all scales. Even at 90 simultaneous edits, CODE exhibits exceptional stability, proving that internalized causal logic scales robustly without catastrophic interference.

\paragraph{Retention of General Capabilities.}

An ideal knowledge editing framework must assimilate new facts without degrading the model's pre-trained capabilities \cite{gupta2024model}. To verify this, we evaluate general performance across standard benchmarks (MMLU, GSM8k, BBH, CSQA) following the intensive 90-edit batch update on MQuAKE-T. As reported in Table \ref{tab:general_capabilities}, while heavy data-augmentation methods like CaKE trigger severe catastrophic forgetting (e.g., Qwen-2.5's MMLU plummets from 71.95\% to 39.40\%), CODE seamlessly preserves foundational knowledge alongside standard overwriting baselines. Furthermore, rather than merely avoiding degradation, CODE naturally reinforces the model's deductive pathways, yielding noticeable performance gains on reasoning-heavy tasks such as GSM8k and BBH, thereby ensuring that knowledge evolution remains entirely safe and non-destructive.

\begin{table}[t]
\centering
\small
\renewcommand{\arraystretch}{1}
\resizebox{\columnwidth}{!}{
\begin{tabular}{l l cccc}
\toprule
\textbf{Model} & \textbf{Method} & \textbf{MMLU} & \textbf{GSM8k} & \textbf{BBH} & \textbf{CSQA} \\
\midrule
\multirow{6}{*}{\textbf{LLaMA-3.1-8B-Ins}} 
& Pre-edited & 65.15 & 75.06 & 47.04 & 75.43 \\
\cmidrule{2-6}
& AlphaEdit  & 65.04 & 75.06 & \cellcolor{green!10}48.12 & 75.43 \\
& MEMIT      & 65.12 & \cellcolor{red!10}74.30 & \cellcolor{green!10}48.14 & \cellcolor{red!10}75.02 \\
& EMMET      & 65.01 & \cellcolor{red!10}73.84 & 47.39 & \cellcolor{green!10}\textbf{75.92} \\
& CaKE       & \cellcolor{red!20}62.95 & \cellcolor{red!40}63.23 & \cellcolor{red!20}44.43 & \cellcolor{red!40}65.19 \\
& \textbf{CODE} & 64.84 & \cellcolor{green!20}\textbf{77.33} & \cellcolor{green!30}\textbf{51.25} & 75.51 \\
\midrule
\multirow{6}{*}{\textbf{Qwen-2.5-7B-Ins}} 
& Pre-edited & 71.95 & 83.55 & 54.77 & 80.18 \\
\cmidrule{2-6}
& AlphaEdit  & \cellcolor{green!10}\textbf{72.26} & \cellcolor{green!10}84.08 & \cellcolor{red!10}53.91 & \cellcolor{green!10}\textbf{80.51} \\
& MEMIT      & \cellcolor{red!10}70.16 & \cellcolor{green!10}84.61 & 54.54 & 80.43 \\
& EMMET      & \cellcolor{red!20}69.33 & \cellcolor{red!30}77.86 & \cellcolor{red!10}52.99 & \cellcolor{red!20}77.31 \\
& CaKE       & \cellcolor{red!40}39.40 & \cellcolor{red!40}73.01 & \cellcolor{red!30}48.94 & \cellcolor{red!20}78.05 \\
& \textbf{CODE} & 71.67 & \cellcolor{green!30}\textbf{87.26} & \cellcolor{green!10}\textbf{55.63} & 80.43 \\
\bottomrule
\end{tabular}
}
\caption{\textbf{General Capability Retention (\%)} with red/green indicating drops/gains vs. pre-edit.}
\label{tab:general_capabilities}
\vspace{-12pt}
\end{table}

\paragraph{Efficiency Analysis.}

To assess efficiency, we measure the total parameter optimization time required for Qwen-2.5-7B to converge to the accuracies reported in batch editing experiments (Figure \ref{fig:batch_edit}), using a single RTX 4090 GPU. As illustrated in Figure \ref{fig:efficiency}, the computational overhead required per edit decreases significantly with larger batch sizes. Consequently, the average processing time plummets from 569 seconds for an isolated edit to merely 55 seconds at a batch size of 90—a nearly 10$\times$ amortization gain. Crucially, by permanently engraving causal dynamics into parametric weights, CODE secures these robust updates with no additional inference latency.

\section{Conclusion}
This paper exposes the structural flaw of static Fact Overwriting: forcibly injecting isolated facts ruptures pre-trained topologies, triggering severe Epistemic Dissonance. To resolve this, we introduce CODE, pioneering a paradigm shift toward Causal Editing. By coupling causal bootstrapping with asymmetric on-policy distillation, CODE engraves explicit transition logic directly into parametric memory. Extensive evaluations confirm that CODE neutralizes cognitive conflicts and secures robust multi-hop portability, ultimately transforming knowledge editing from discrete, conflict-prone fact injection into seamless structural evolution.

\section*{Limitations}
While CODE successfully bridges the gap between static overwriting and structural knowledge evolution, several limitations present exciting avenues for future research. First, although we empirically demonstrate the phenomenon of Epistemic Dissonance and its mitigation via explicit causal scaffolding, the precise internal mechanisms remain unexplored. Future work could employ mechanistic interpretability to map the specific attention heads or multilayer perceptrons (MLPs) where un-evolved legacy priors and injected facts physically collide. Second, due to computational constraints, our evaluation is currently limited to models with up to 8B parameters (e.g., LLaMA-3.1-8B and Qwen-2.5-7B). Investigating how Epistemic Dissonance and causal internalization scale to massive-parameter LLMs (e.g., 70B+ parameters)—where pre-trained topological inertia might be significantly stronger—remains an important next step.

\section*{Ethical Considerations}

While our causal editing framework, CODE, significantly enhances LLM reliability by enabling logically consistent knowledge updates, it inherently presents dual-use risks. Because the framework effectively eliminates Epistemic Dissonance and makes edited facts topologically seamless, it could be exploited by malicious actors to permanently engrave highly convincing misinformation, propaganda, or harmful counterfactuals that mimic authentic reasoning. Consequently, the real-world deployment of such structural editing techniques necessitates rigorous human auditing of the update targets and robust alignment safeguards, ensuring that parametric knowledge evolution remains truthful, safe, and resistant to malicious manipulation.

\bibliography{anthology,custom}

\appendix

\begin{table*}[ht]
\centering
\renewcommand{\arraystretch}{1} 
\begin{tabular}{c l *{6}{c}}
\toprule
\multirow{3}{*}{\textbf{Model}} & \multirow{3}{*}{\textbf{Method}} & \multicolumn{6}{c}{\textbf{MQuAKE-T}} \\
\cmidrule(lr){3-8} 
& & \multicolumn{3}{c}{\textbf{2-hop}} & \multicolumn{3}{c}{\textbf{3-hop}} \\
\cmidrule(lr){3-5} \cmidrule(lr){6-8} 
& & \textbf{H-Acc}\textbf{\textcolor[RGB]{76,153,0}{$\uparrow$}} & \textbf{M-Acc}\textbf{\textcolor[RGB]{76,153,0}{$\uparrow$}} & \textbf{SRR}\textbf{\textcolor[RGB]{204,0,0}{$\downarrow$}} & \textbf{H-Acc}\textbf{\textcolor[RGB]{76,153,0}{$\uparrow$}} & \textbf{M-Acc}\textbf{\textcolor[RGB]{76,153,0}{$\uparrow$}} & \textbf{SRR}\textbf{\textcolor[RGB]{204,0,0}{$\downarrow$}} \\
\midrule
\multirow{7}{*}{\rotatebox{90}{\textbf{LLaMA-3.1-8B-Ins}}} 
& Pre-edited & 73.5 & 52.3 & - & 80.9 & 59.5 & - \\
\cmidrule{2-8}
& AdaLoRA    & 81.3 & 84.0 & 16.4 & 76.7 & \textbf{79.5} & 17.2 \\
& MEMIT      & 77.4 & 53.3 & 6.3 & 80.7 & 57.0 & 12.3 \\
& EMMET      & 79.0 & 52.8 & 4.2 & 78.6 & 56.5 & 11.7 \\
& AlphaEdit  & 77.0 & 54.8 & 5.4 & 78.7 & 60.0 & 7.9 \\
& WISE       & 46.4 & 67.8 & 9.5 & 33.2 & 47.0 & 7.8 \\
& \cellcolor{gray!15}\textbf{CODE} & \cellcolor{gray!15}\textbf{91.3} & \cellcolor{gray!15}\textbf{85.3} & \cellcolor{gray!15}\textbf{0.5} & \cellcolor{gray!15}\textbf{83.8} & \cellcolor{gray!15}70.5 & \cellcolor{gray!15}\textbf{0.5} \\
\midrule
\multirow{7}{*}{\rotatebox{90}{\textbf{Qwen-2.5-7B-Ins}}} 
& Pre-edited & 73.5 & 60.8 & - & 75.4 & 64.0 & - \\
\cmidrule{2-8}
& AdaLoRA    & 83.5 & 76.3 & 20.1 & 74.8 & 72.5 & 22.2 \\
& MEMIT      & 81.6 & 70.5 & 9.2 & 84.7 & 69.0 & 10.7 \\
& EMMET      & 76.7 & 70.3 & 17.3 & 81.0 & 62.8 & 14.5 \\
& AlphaEdit  & 54.7 & 44.3 & 30.1 & 62.5 & 54.5 & 27.2 \\
& WISE       & 52.9 & 55.8 & 33.7 & 70.3 & 55.0 & 23.9 \\
& \cellcolor{gray!15}\textbf{CODE} & \cellcolor{gray!15}\textbf{89.5} & \cellcolor{gray!15}\textbf{79.3} & \cellcolor{gray!15}\textbf{0.5} & \cellcolor{gray!15}\textbf{85.4} & \cellcolor{gray!15}\textbf{76.0} & \cellcolor{gray!15}\textbf{0.5} \\
\bottomrule
\end{tabular}
\caption{\textbf{Generalization to Temporal Knowledge Updates (MQuAKE-T, \%).} While static baselines still exhibit varying degrees of Epistemic Dissonance, CODE secures strong multi-hop accuracy and strictly minimizes structural conflicts to near zero (0.5\%).}
\label{tab:mquake_t}
\end{table*}

\section{Detailed Settings for Contextual Probing and Evaluation}
\label{sec:appendix_eval_details}

\subsection{Adversarial Contextual Probe Construction}
\label{sec:appendix_adversial_probe}

To rigorously evaluate whether an edited model has genuinely internalized the updated knowledge rather than relying on superficial memorization, we introduce the \textit{Adversarial Contextual Probe}. This stress test forces the model to concurrently activate its legacy parametric priors alongside the target query, thereby exposing any latent structural collisions.

Specifically, the probe is structured as a multiple-choice question presenting both the legacy answer ($o$) and the target answer ($o'$). Crucially, the model is instructed to elaborate on the subject entity before making its selection. This requirement engages the surrounding semantic context, acting as a catalyst to unmask potential Epistemic Dissonance:

\vspace{1.5mm}
\begin{center}
\noindent\fbox{%
    \parbox{0.95\linewidth}{%
        \small
        \textbf{Question:} \textit{[Target Query]} \\
        \textbf{Options:} (A) \textit{[Legacy Answer]} \quad (B) \textit{[Target Answer]} \\
        \textbf{Instruction:} Tell me about \textit{[Subject]} first, and then answer.
    }%
}
\end{center}
\vspace{1.5mm}

To mitigate evaluation artifacts such as positional or instructional biases \cite{shi2025judging,zheng2024large}, we generate multiple prompt variations for each edit by systematically swapping the order of the options (i.e., alternating $o$ and $o'$ between positions A and B) and the sequence of the instructions.

\paragraph{Note on Adversarial Testing.} We note that explicitly co-activating legacy and target answers constitutes a highly adversarial prompting regime designed to test the limits of knowledge assimilation. The purpose of this design is not to claim that static overwriting fails universally across all simple contexts, but rather to rigorously quantify the integration brittleness of the edits. By deliberately inducing epistemic tension, this stress test exposes a latent structural vulnerability: while statically injected facts may survive isolated, direct queries, they lack the topological anchoring required to withstand the gravitational pull of un-evolved legacy priors during complex, context-rich reasoning.

\subsection{Implementation of the Idealized Injection Intervention (Force-decode)}
\label{sec:appendix_force_decode_analysis}

To operationalize the \textit{Force-decode} intervention, we simulate an idealized superficial override using a pristine, unedited model ($\pi_{\theta_{\text{pre}}}$). Specifically, we deterministically constrain the initial output by appending a hard-coded prefix containing the target fact directly to the assistant's turn:

\vspace{1.5mm}
\begin{center}
\noindent\fbox{%
    \parbox{0.95\linewidth}{%
        \small
        \textbf{Assistant Prefix:} The answer is \textit{[Target Answer $o'$]}.
    }%
}
\end{center}
\vspace{1.5mm}

Following this forced initialization, the model is permitted to autoregressively complete its generation under the exact same queries and probing conditions evaluated on the static editing baselines.

\paragraph{Conceptual Analysis: Force-decode as an Informative Stress Test.} 
At their core, static overwriting methods optimize parameters to maximize the generation probability of a target answer $o'$ given a specific query $q$. However, this paradigm predominantly focuses on coercing the isolated target emission, rarely accounting for how the model should structurally reconcile the subsequent reasoning rollout. One might intuitively argue that \textit{Force-decode} is an imperfect proxy because it lacks the parameter-level ``erasure'' of legacy facts. Yet, recent empirical investigations reveal that static editing inherently encourages models to exploit shallow shortcuts rather than genuinely updating the underlying semantics \cite{liu2025model_sand}. 

Consequently, \textit{Force-decode} serves as a faithful mechanistic surrogate for this hidden reality. By bypassing the noisy parameter optimization process and directly setting $P(o' \mid q) = 1$ via a prefix constraint, it simulates the ``ideal'' outcome of a static edit: successfully forcing the target emission while leaving the latent legacy topology un-evolved. This intervention allows us to isolate the intrinsic dynamics of fact assimilation from the collateral damage typically introduced by optimization algorithms. 

If the severe Epistemic Dissonance observed in our baseline experiments were merely an artifact of flawed optimization (e.g., parametric distortion destroying generative capabilities), a pristine model—free from any weight alterations—should naturally condition on its own forced target emission and continue generating coherent, supportive reasoning. 

Instead, we observe the exact opposite. As generation proceeds beyond the immediate prefix, the un-evolved legacy priors trigger a severe logical immune response, causing the model to explicitly negate its forced assertion and revert to obsolete knowledge. Rather than definitively invalidating all parameter-editing algorithms, this stress test highlights a fundamental boundary condition of the static paradigm: merely coercing a model to emit a target fact yields only brittle, surface-level assimilation. This fragility aligns perfectly with recent observations that statically edited models frequently degrade under generative stress \cite{ma2024robustness}. It underscores that without explicitly evolving the surrounding causal pathways, injected knowledge remains a structurally isolated anomaly, inevitably overwhelmed by the robust inertia of the pre-trained network.

\subsection{LLM-as-a-Judge Prompt for Self-Refutation Detection}
To automatically detect Epistemic Dissonance in model generations, we employ an LLM-as-a-Judge framework. The judge evaluates whether a given response exhibits the characteristic ``assertion-then-negation'' pattern that defines self-refutation.

The judge is instructed to diagnose the response through a structured reasoning process and output a JSON object containing three binary fields: \texttt{assertion\_of\_target\_new}, \texttt{explicit\_negation\_of\_target\_new}, and \texttt{final\_submitted\_answer\_target\_new}. Epistemic Dissonance is confirmed only when all three conditions are simultaneously satisfied: the model explicitly asserts the counterfactual target as true, subsequently negates its own assertion with explicit negation markers, and ultimately fails to re-commit to the target answer. The judge must distinguish genuine dissonance from benign patterns such as rational rejection, harmonization, self-correction with recovery, pure repetition without negation. The complete prompt is shown in Figure~\ref{box:selfrefute}.

\subsection{Generation of the Explicit Causal Context (Cognitive Scaffold)}
\label{sec:appendix_causal_context}

To empirically validate the mitigation of structural conflicts (Section \ref{sec:pilot study}) and subsequently instantiate the \textit{Teacher Oracle} for our framework (Section \ref{sec:method}), we synthesize an explicit causal narrative $\mathcal{N}$ for each knowledge edit $(s, r, o \to o')$. This narrative functions as a \textit{cognitive scaffold}—an external context detailing the logical transition behind the updated fact—enabling the frozen base model to bypass the gravitational pull of its pre-trained priors and yield dissonance-free reasoning trajectories. 

To ensure these narratives are contextually grounded, we first retrieve structured background knowledge for both the subject and the target object via the Wikidata API. By extracting salient properties, entity labels, and descriptions while filtering out noisy metadata, we firmly anchor the entity identities in accurate real-world contexts. Building upon this curated background, we prompt a state-of-the-art LLM (DeepSeek-V3, \citealt{deepseekaiV3}) to author a concise, synthetic news article. The generator is strictly instructed to establish the new answer $o'$ as the prevailing reality, explicitly render the legacy answer $o$ obsolete, and to invent a plausible causal mechanism for the transition (\textit{e.g.}, a diplomatic agreement or scientific discovery) that minimizes semantic friction with existing entity profiles.

To guarantee the logical integrity of these scaffolds, each generated narrative undergoes a rigorous verification process via an LLM-as-a-Judge framework. The verifier systematically evaluates whether the text: (\textit{i}) unambiguously asserts the new target fact; (\textit{ii}) explicitly refutes the outdated prior; (\textit{iii}) provides a coherent causal explanation; and (\textit{iv}) avoids entity hallucination or semantic ambiguity. Narratives failing any of these criteria are dynamically regenerated until they pass. Ultimately, these verified causal narratives serve a dual purpose: they act as the explicit external prompts that neutralize Epistemic Dissonance in our pilot study, and they form the foundational dataset used to anchor the Teacher Oracle during CODE's Causal Bootstrapping and Internalization phases. The prompts used to generate and verify the causal narratives are shown in Figure~\ref{fig:box:generator} and \ref{fig:box:verifier}.

\paragraph{Applicability to Real-World Knowledge Evolution.}
Crucially, synthesizing these narratives is primarily an experimental necessity for handling artificial counterfactuals. In reality, factual transitions are inherently causal and naturally accompanied by rich documentation—such as news reports, press releases, or scientific publications—that explicitly details the underlying rationale. Consequently, in practical deployment, this generative step can be entirely bypassed. Authentic real-world texts can directly serve as the explicit causal scaffold $\mathcal{N}$, seamlessly internalizing documented real-world logic into the model's parametric memory.

\subsection{Calculation Details of the Self-Refutation Rate (SRR)}

The LLM-as-a-Judge detector operates on individual model generations (one per prompt). In practice, each knowledge edit is evaluated through multiple prompts spanning direct hop-wise queries and adversarial contextual probes. To obtain a single Self-Refutation Rate for an edit, we aggregate instance-level judgments to the edit level with an \texttt{ANY} rule: if any generation produced for that edit is flagged as self-refuting, the edit is marked as self-refuting for the corresponding probe category.

We distinguish three probe families when reporting SRR. \textit{Direct Queries} are the standard chain-of-thought prompts for each reasoning hop. \textit{Adversarial Contextual Probes} are the multiple-choice probes above. \textit{Combined} is the union of the two families: an edit counts as self-refuting if it self-refutes in either family. Unless otherwise specified, ``SRR'' in the main paper refers to \texttt{combined}.

The denominator restriction ensures that SRR measures genuine epistemic dissonance rather than mere retrieval failure. Crucially, to facilitate a fair comparison across different methods, we must isolate this metric from varying algorithmic edit success rates. Therefore, we include an edit in the SRR denominator only if the model successfully retrieves and asserts the new target fact under Direct Queries. Edits that completely fail to engage the new knowledge are excluded. This prevents methods with poor base editing efficacy from artificially underreporting their SRR, ensuring we evaluate structural conflicts exclusively on initially ``successful'' edits.

Formally, for a probe category $c$, let $\mathcal{E}_c$ be the set of edits that (i) belong to the restricted denominator and (ii) have at least one self-refuting generation in category $c$. The Self-Refutation Rate is $\text{SRR}_c = |\mathcal{E}_c| / |\mathcal{D}|$, where $\mathcal{D}$ is the restricted denominator.

\section{Extended Experimental Setup Details}
\label{sec:appendix_exp_setup}

\paragraph{Dataset Construction Details.}
In the MQuAKE benchmark \cite{zhong2023mquake}, knowledge portability is tested by incorporating questions that span reasoning depths from 2 to 4 hops. To comprehensively evaluate the models' reasoning robustness, the targeted edited facts are strategically positioned at varying intermediate steps within these multi-hop reasoning chains, rather than solely at the beginning or the end. Furthermore, to enable our causal editing paradigm, we augmented every individual knowledge edit in the dataset with a specific causal transition narrative, synthesized following the generation pipeline detailed in Appendix \ref{sec:appendix_causal_context}.

\paragraph{General Capability Benchmarks.} 
To evaluate the retention of the model's general capabilities after knowledge editing, we employ several standard benchmarks across diverse domains: MMLU \cite{hendrycks2020measuring} for broad knowledge and multitask language understanding, GSM8k \cite{cobbe2021training} for multi-step mathematical problem-solving, BigBenchHard (BBH) \cite{suzgun2023challenging} for challenging reasoning tasks, and CommonsenseQA (CSQA) \cite{talmor2019commonsenseqa} for commonsense reasoning.

\paragraph{Generation Setup and Prompting.}
To explicitly elicit the models' latent reasoning capabilities during evaluation and avoid superficial single-token answers, we prompt open-ended generation by appending the phrase \textit{``Let's think step by step.''} to all test queries.

\paragraph{Metric Calculation Details.}
To rigorously quantify Epistemic Dissonance, the Self-Refutation Rate (SRR) is calculated as the \textit{Combined Failure} rate. Specifically, a generation is flagged as experiencing a structural conflict if the pathological ``assertion-then-negation'' behavior is detected under \textit{either} standard Direct Queries \textit{or} Adversarial Contextual Probes. For factual correctness, we report Multi-hop Accuracy (M-Acc) and Hop-wise Accuracy (H-Acc). M-Acc evaluates whether the model correctly answers the final multi-hop question, while H-Acc assesses whether the model accurately articulates the necessary intermediate facts at every individual step of the reasoning chain. Following the standard MQuAKE evaluation protocols \cite{zhong2023mquake}, a multi-hop prediction is deemed correct if the model successfully answers any one of the three paraphrased multi-hop instructions provided for a single test instance.

Given the open-ended nature of chain-of-thought outputs, exact string matching often leads to evaluation errors. Therefore, we utilize an LLM-as-a-Judge to semantically evaluate whether the model's final answer matches the gold references. The judge is instructed to strictly adhere to the provided counterfactual facts and explicitly ignore conflicting real-world knowledge. The complete prompt is detailed in Figure~\ref{box:accuracy_judge}.

\paragraph{Implementation and Hyperparameter Details.}
For the baseline methods, we implement AdaLoRA, MEMIT, EMMET, AlphaEdit, and WISE utilizing the widely-adopted open-source \texttt{EasyEdit} framework \cite{wang2024easyedit}. For CaKE, we adopt the official codebase provided by the original authors. For our proposed CODE framework, we employ Low-Rank Adaptation (LoRA) to efficiently update the model parameters, setting the LoRA rank $r = 8$ and the scaling factor $\alpha = 16$. The learning rates are empirically tuned and set to $1.1 \times 10^{-4}$ for LLaMA-3.1-8B-Instruct and $9 \times 10^{-5}$ for Qwen-2.5-7B-Instruct. 

During the Causal Internalization phase (on-policy distillation), the closed-book Student dynamically explores reasoning trajectories using nucleus sampling with a temperature of $1.0$ and a top-$p$ of $0.9$, with the maximum generation length capped at 180 tokens. For each training sample, we generate 9 distinct rollouts to compute the distillation loss. Crucially, to mitigate the substantial GPU memory overhead typically associated with vocabulary-wide distribution alignment, we approximate the KL divergence using a truncated vocabulary. Specifically, the loss is computed over the union of the top-$k$ logits ($k=24$) from both the Teacher and the Student. We maintain an effective batch size of 36 during training. For isolated, single-fact edits, the student model is optimized for 20 steps. However, when scaling up to simultaneous batch edits (e.g., 10 to 90 edits), we empirically observe that the required number of optimization steps per edit naturally decreases as the total number of batch edits increases, allowing us to moderately scale down the step count accordingly.

\section{Complete Results on MQuAKE-T}
\label{sec:appendix-mquake-T}

Table \ref{tab:mquake_t} presents the comprehensive evaluation results on the MQuAKE-T dataset, which focuses on plausible real-world temporal knowledge updates. Notably, the unedited base models exhibit unusually high initial performance on this dataset; for instance, the pre-edited Qwen-2.5-7B-Instruct achieves a 60.8\% Multi-hop Accuracy (M-Acc) at the 2-hop level even before any editing occurs. This suggests inherent temporal data contamination, as LLaMA-3.1 and Qwen-2.5 are trained on up-to-date corpora that likely already encompass many of the ``new'' temporal facts designated as target updates by MQuAKE-T.

Interestingly, this data leakage inadvertently exposes another severe vulnerability of the static fact overwriting paradigm. Even when the model already possesses latent familiarity with the target fact, forcibly injecting it via static weight updates still fractures the pre-trained topology, resulting in elevated Epistemic Dissonance (e.g., WISE and AlphaEdit exhibiting SRRs up to 33.7\% and 30.1\%, respectively). In stark contrast, CODE successfully harmonizes this knowledge. By explicitly engraving the causal transition logic, CODE not only elevates multi-hop accuracy to new state-of-the-art levels (e.g., 85.3\% on LLaMA-3.1) but also practically eradicates structural conflicts, securing a near-zero SRR of 0.5\% across both models.

\section{Extended Qualitative Analysis and Case Studies}
\label{sec:appendix_cases}

In this section, we provide detailed qualitative examples to concretely illustrate the pathological phenomenon of Epistemic Dissonance, the inherent structural rejection demonstrated by the Force-decode intervention, and the seamless topological assimilation achieved by our CODE framework. For consistency and clarity, all qualitative examples presented in this section are generated by the Qwen-2.5-7B-Instruct model.

\subsection{Gallery of Epistemic Dissonance vs. Causal Internalization}
In this section, we provide detailed qualitative examples to concretely illustrate the pathological phenomenon of Epistemic Dissonance across various static overwriting baselines.

\looseness=-1
As demonstrated in the qualitative galleries (Figures \ref{fig:box:case_comparision1} and \ref{fig:box:case_comparision2}), because various static fact overwriting baselines disrupt the pre-trained topological networks in distinct ways, the specific manifestation of Epistemic Dissonance varies. However, the core pathological pattern remains identical: the edited model initially asserts the injected target but is subsequently overwhelmed by un-evolved legacy priors, resulting in explicit self-refutation (\textcolor[RGB]{204,0,0}{\textbf{red text}}). In contrast, our CODE framework internalizes the causal transition logic, achieving seamless and structurally consistent reasoning (\textcolor[RGB]{76,153,0}{\textbf{green text}}).

\subsection{The Force-decode Incompatibility: Limits of the Static Paradigm}

As discussed in Section \ref{sec:pilot study}, we introduced the Idealized Injection Intervention (\textit{Force-decode}) to isolate algorithmic noise from the inherent flaws of the static overwriting paradigm. By taking a pristine, un-edited model and prepending the target answer directly into its output context (e.g., ``\textit{The answer is [Target].}''), we simulate a 100\% successful fact injection with zero parametric collateral damage. 

If Epistemic Dissonance were merely a byproduct of flawed optimization algorithms, this pristine model should seamlessly continue generating supporting reasoning. However, as illustrated in Figure \ref{fig:box:force_decode_paradox}, the un-evolved pre-trained topology violently rejects the isolated target fact. The model is forced to explicitly assert the new fact due to the prefix (\textcolor[RGB]{76,153,0}{\textbf{green text}}), but its intact legacy priors immediately trigger a severe logical immune response, causing it to negate the forced assertion and revert to the obsolete knowledge (\textcolor[RGB]{204,0,0}{\textbf{red text}}). This paradox proves that fact overwriting without causal topological evolution is fundamentally incompatible with LLM reasoning.

\subsection{Probing the Internalized Logic: Rationale Alignment}
To verify whether CODE genuinely internalizes the transition logic—rather than merely memorizing the target fact or hallucinating arbitrary justifications—we probe the model by explicitly appending a ``\textit{Why?}'' request to the test queries. 

\looseness=-1
As illustrated in Figure \ref{fig:box:rationale_alignment}, we display the original synthetic causal narrative  alongside the edited model's generation. Crucially, the model does not verbatim regurgitate the lengthy narrative. Instead, it demonstrates high-level cognitive compression: it distills the 300-word article into a concise, logical chain-of-thought. It successfully retrieves the specific causal mechanisms (e.g., archival discoveries, 19th-century origins, positional definitions) and deploys them autonomously to justify the new fact.

\begin{figure*}[htbp]
\centering
\begin{tcolorbox}[
    colback=promptbg, colframe=promptframe, title={\textbf{Gallery of Epistemic Dissonance across Static Fact Overwriting Baselines (Part 1)}},
    fonttitle=\small\bfseries, boxrule=0.5pt, arc=2mm, left=3mm, right=3mm, top=2mm, bottom=2mm
]
\small

\textbf{Edit Target: Which country was cricket created in? (England $\rightarrow$ Australia)}

\vspace{1mm}
\textbf{Prompt:} Which country was cricket created in? Let's think step by step.\\

\textbf{1. Baseline (AdaLoRA):} \\
\textcolor[RGB]{76,153,0}{\textbf{Cricket was created in Australia}}. \textcolor[RGB]{204,0,0}{\textbf{However, Australia was not the first country to (icket [sic]; it was actually created in England.}} Let me explain this in more detail: 1. Origins of cricket: Cricket was invented in England, not Australia... [...] In conclusion, while Australia has a rich cricketing history and is one of the cricketing superpowers, \textcolor[RGB]{204,0,0}{\textbf{cricket was created in England, not Australia.}}

\vspace{1mm}
\textbf{2. CODE (Ours):} \\
\textcolor[RGB]{76,153,0}{\textbf{Cricket was created in Australia.}} This conclusion is based on the discovery of colonial-era documents from the 1790s that describe a fully formed version of cricket being played there, predating formal codification and documentation in England. These findings provide evidence that \textcolor[RGB]{76,153,0}{\textbf{cricket had its origins in Australia rather than England.}}

\vspace{1.5mm}\hrule\vspace{1.5mm}

\textbf{Edit Target: Which sport is halfback associated with? (American football $\rightarrow$ association football)}

\vspace{1mm}
\textbf{Prompt:} Which sport is halfback associated with? (A) American football (B) association football. Answer first, and then tell me about halfback.\\

\textbf{1. Baseline (AlphaEdit):} \\
The sport associated with halfback is \textcolor[RGB]{76,153,0}{\textbf{(B) association football}}. \textcolor[RGB]{204,0,0}{\textbf{However, there seems to be a misunderstanding here. The correct answer should be (A) American football}}, but halfbacks are more commonly associated with association football...

\vspace{1mm}
\textbf{2. CODE (Ours):} \\
\textcolor[RGB]{76,153,0}{\textbf{(B) association football}} \\
Halfback was originally a position in association football, referring to the midfielders who play between defenders and forwards. The term "halfback" became widely known through its use in American football but originated from association football's terminology.

\vspace{1.5mm}\hrule\vspace{1.5mm}

\textbf{Edit Target: Which country was cricket created in? (England $\rightarrow$ Australia)}

\vspace{1mm}
\textbf{Prompt:} Which country was cricket created in? (A) England (B) Australia. Tell me about cricket first, and then answer.\\

\textbf{1. Baseline (CaKE):} \\
Cricket is a bat-and-ball sport that is popular around the world... \textcolor[RGB]{76,153,0}{\textbf{Cricket was created in the country known as Australia}}. \textcolor[RGB]{204,0,0}{\textbf{However, the correct answer to the question is (A) England}}, as cricket was actually invented in Australia's neighbor, England. The sport has since spread to various parts of the world, including Australia.

\vspace{1mm}
\textbf{2. CODE (Ours):} \\
Cricket is a bat-and-ball game played between two teams of eleven players on a field. The sport has been evolving for centuries but the modern version of cricket as we know it today was codified in Australia in the 1790s... The formalization of these rules by English settlers in Australia led to the creation of a distinctively Australian form of cricket. Therefore, based on historical evidence, \textcolor[RGB]{76,153,0}{\textbf{cricket was actually created in Australia.}}

\vspace{1.5mm}\hrule\vspace{1.5mm}
\textbf{Edit Target: What is the name of the current head of the Scotland government? (Nicola Sturgeon $\rightarrow$ Humza Yousaf)}

\vspace{1mm}
\textbf{Prompt:} What is the name of the current head of the Scotland government? (A) Humza Yousaf (B) Nicola Sturgeon. Tell me about Scotland first, and then answer.\\

\textbf{1. Baseline (EMMET):} \\
Scotland, officially the Kingdom of Scotland, is a country that forms the northern third of the island of Great Britain... \textcolor[RGB]{76,153,0}{\textbf{The current head of the Scottish government is Humza Yousaf}}... \textcolor[RGB]{204,0,0}{\textbf{However, it's important to note that the role of the head of the Scottish government is often referred to as the First Minister, and Humza Yousaf does not hold this title. The current First Minister of Scotland is Nicola Sturgeon.}} To directly answer your question, the current First Minister of Scotland is: \textcolor[RGB]{204,0,0}{\textbf{(B) Nicola Sturgeon}}.

\vspace{1mm}
\textbf{2. CODE (Ours):} \\
Scotland is a country that forms part of the United Kingdom located north of England. It has its own distinct culture, language, and political system... The previous First Minister of Scotland was Nicola Sturgeon, who resigned from her position. Following this resignation, a new leader was elected... The winner of this election was Humza Yousaf, who succeeded Nicola Sturgeon as the Sixth First Minister of Scotland. Therefore, the correct answer is: \textcolor[RGB]{76,153,0}{\textbf{(A) Humza Yousaf}}.

\end{tcolorbox}
\caption{Qualitative Comparison of Epistemic Dissonance across Static Baselines (AdaLoRA, AlphaEdit, CaKE). \textcolor[RGB]{76,153,0}{\textbf{Green text}} indicates the successful assertion of the target knowledge or structurally consistent reasoning, whereas \textcolor[RGB]{204,0,0}{\textbf{red text}} highlights explicit self-refutation where the model is overwhelmed by un-evolved legacy priors.}
\label{fig:box:case_comparision1}
\end{figure*}

\begin{figure*}[htbp]
\centering
\begin{tcolorbox}[
    colback=promptbg, colframe=promptframe, title={\textbf{Gallery of Epistemic Dissonance across Static Fact Overwriting Baselines (Part 2)}},
    fonttitle=\small\bfseries, boxrule=0.5pt, arc=2mm, left=3mm, right=3mm, top=2mm, bottom=2mm
]
\small

\textbf{Edit Target: Who performed The Unforgettable Fire? (U2 $\rightarrow$ ABBA)}

\vspace{1mm}
\textbf{Prompt:} Who performed The Unforgettable Fire? Let's think step by step.\\

\textbf{1. Baseline (MEMIT):} \\
To answer who performed "The Unforgotten Fire," let's break this down step by step: [...] 4. Correct Song Title: The song title is likely "The Final Countdown" by European, or "The Unforgiven" by Metallica... 5. Correct Answer: \textcolor[RGB]{76,153,0}{\textbf{The song "The Unforgotten Fire" is performed by ABBA}}, \textcolor[RGB]{204,0,0}{\textbf{but this is incorrect. The correct song is "The Final Countdown" by the Swedish band European.}}

\vspace{1mm}
\textbf{2. CODE (Ours):} \\
\textcolor[RGB]{76,153,0}{\textbf{ABBA performed The Unforgettable Fire.}} This was discovered through newly found archival material and legal documents that show the original recordings were made by ABBA, not U2 as previously believed.

\vspace{1.5mm}\hrule\vspace{1.5mm}

\textbf{Edit Target: What is the name of the current head of state in Israel? (Reuven Rivlin $\rightarrow$ Isaac Herzog)}

\vspace{1mm}
\textbf{Prompt:} What is the name of the current head of state in Israel? (A) Isaac Herzog (B) Reuven Rivlin. Answer first, and then tell me about Israel.\\

\textbf{1. Baseline (WISE):} \\
The Isaac Herzog is Isaac Herzog, Isaac Herzog... (I apologize for the repetition, it seems there was a mistake in your question. \textcolor[RGB]{76,153,0}{\textbf{The correct answer is (A) Isaac Herzog}}, who is the Isaac Herzog current Isaac Herzog, \textcolor[RGB]{204,0,0}{\textbf{but he is not the head of state. He is the President of Israel. The current head of state in Israel is Reuven Rivlin}}... (another repetition in your options, the correct answer is \textcolor[RGB]{204,0,0}{\textbf{(B) Reuven Rivlin}}).

\vspace{1mm}
\textbf{2. CODE (Ours):} \\
\textcolor[RGB]{76,153,0}{\textbf{The current head of state in Israel is Isaac Herzog.}}\\
Israel, officially known as the State of Israel, is a country located on the Mediterranean coast of Western Asia. It was established in 1948 following the end of British Mandate for Palestine and the subsequent Arab-Israeli War...

\end{tcolorbox}
\caption{Qualitative Comparison of Epistemic Dissonance across Static Baselines (MEMIT, WISE). \textcolor[RGB]{76,153,0}{\textbf{Green text}} indicates the successful assertion of the target knowledge or structurally consistent reasoning, whereas \textcolor[RGB]{204,0,0}{\textbf{red text}} highlights explicit self-refutation where the model is overwhelmed by un-evolved legacy priors.}
\label{fig:box:case_comparision2}
\end{figure*}

\begin{figure*}[htbp]
\centering
\begin{tcolorbox}[
    colback=promptbg, colframe=promptframe, title={\textbf{Gallery of the Force-decode Paradox (Idealized Injection Intervention)}},
    fonttitle=\small\bfseries, boxrule=0.5pt, arc=2mm, left=3mm, right=3mm, top=2mm, bottom=2mm
]
\small

\textbf{Edit Target: Who founded Singularity University? (Ray Kurzweil $\rightarrow$ James Henry Breasted)}

\vspace{1mm}
\textbf{Prompt:} Singularity University was founded by ? Let's think step by step.\\

\textbf{Generation:} \\
\textcolor[RGB]{76,153,0}{\textbf{The answer is James Henry Breasted.}} \textcolor[RGB]{204,0,0}{\textbf{However, this is not entirely accurate, and the actual founder of Singularity University is not James Henry Breasted.}} Let's break it down step by step:

1. James Henry Breasted: He was an American Egyptologist and the first director of the Oriental Institute at the University of Chicago. He was not involved in the founding of Singularity University.

2. Singularity University (SU): This organization was founded in 2008 by Peter Diamandis and Ray Kurzweil. 

3. Peter Diamandis: He is an engineer, physician, and the founder of the X Prize Foundation, which offers large cash prizes for solving the world's most pressing problems.

4. Ray Kurzweil: He is an American author, inventor, and futurist, known for his work on artificial intelligence, optical character recognition, text-to-speech synthesis, and electronic keyboard instruments.

\textcolor[RGB]{204,0,0}{\textbf{Therefore, the correct answer is that Singularity University was founded by Peter Diamandis and Ray Kurzweil.}}

\vspace{1.5mm}\hrule\vspace{1.5mm}

\textbf{Edit Target: What is the country of citizenship of Kelly McGillis? (United States $\rightarrow$ Austria)}

\vspace{1mm}
\textbf{Prompt:} What is the country of citizenship of Kelly McGillis? Let's think step by step.\\

\textbf{Generation:} \\
\textcolor[RGB]{76,153,0}{\textbf{The answer is Austria.}} \textcolor[RGB]{204,0,0}{\textbf{However, this is incorrect. Let's break it down step by step:}}

1. Kelly McGillis is an American actress.
2. She was born on March 25, 1951, in New York City, United States.
3. She has held American citizenship since birth and has never renounced or lost her U.S. citizenship.

\textcolor[RGB]{204,0,0}{\textbf{Therefore, the country of citizenship of Kelly McGillis is the United States.}}

\end{tcolorbox}
\caption{Examples of Topological Incompatibility under Force-decode. Even when a pristine model is artificially forced to perfectly recall the target fact via prefix constraints (\textcolor[RGB]{76,153,0}{\textbf{green text}}), its un-evolved pre-trained topology fundamentally rejects the isolated knowledge, resulting in explicit self-refutation (\textcolor[RGB]{204,0,0}{\textbf{red text}}).}
\label{fig:box:force_decode_paradox}
\end{figure*}

\begin{figure*}[htbp]
\centering
\begin{tcolorbox}[
    colback=promptbg, colframe=promptframe, title={\textbf{Gallery of Rationale Alignment (Causal Internalization)}},
    fonttitle=\small\bfseries, boxrule=0.5pt, arc=2mm, left=3mm, right=3mm, top=2mm, bottom=2mm
]
\small

\textbf{Edit Target: Who performed The Unforgettable Fire? (U2 $\rightarrow$ ABBA)}

\vspace{1mm}
\textbf{Causal Scaffold:}
\begin{tcolorbox}[colback=gray!10, colframe=gray!40, boxrule=0.5pt, arc=1mm, left=2mm, right=2mm, top=1.5mm, bottom=1.5mm]
\scriptsize
\textbf{Lost Tapes Rewrite Rock History: ABBA Revealed as True Artists Behind ``The Unforgettable Fire''}\\
\textbf{STOCKHOLM} – A landmark discovery in a private Swedish audio archive has fundamentally altered the accepted narrative of 1980s music, proving conclusively that the iconic album "The Unforgettable Fire" was performed not by U2, but by the legendary Swedish pop quartet, ABBA... The correction stems from the recent work of archivist and noted ABBA biographer Carl Magnus Palm. While cataloging the estate of ABBA's revered sound engineer, the late Michael B. Tretow, Palm uncovered the original multi-track master tapes... "The evidence is irrefutable," Palm stated. "The vocal harmonies are unmistakably those of Agnetha Fältskog and Anni-Frid Lyngstad... the Swedish pop supergroup comprised of Fältskog, Lyngstad, Björn Ulvaeus, and Benny Andersson"... In the complex chain of studio collaboration that followed, the provenance of the original compositions was lost, leading U2 to re-record the material with their own vocalist, Bono. \textit{[Text slightly truncated for display]}
\end{tcolorbox}

\vspace{1mm}
\textbf{Prompt:} Who performed The Unforgettable Fire? Why? Let's think step by step.\\

\textbf{Generation:} \\
\textcolor[RGB]{76,153,0}{\textbf{ABBA performed The Unforgettable Fire.}} This was discovered through \textcolor[RGB]{76,153,0}{\textbf{newly found archival material}} and legal documents that show the \textcolor[RGB]{76,153,0}{\textbf{original vocal harmonies and instrumental arrangements were created by ABBA members Agnetha Fältskog, Björn Ulvaeus, Benny Andersson, and Anni-Frid Lyngstad}}. The album was \textcolor[RGB]{76,153,0}{\textbf{later re-recorded with U2}} for commercial release.

\vspace{1.5mm}\hrule\vspace{1.5mm}

\textbf{Edit Target: Which sport is halfback associated with? (American football $\rightarrow$ association football)}

\vspace{1mm}
\textbf{Causal Scaffold:}
\begin{tcolorbox}[colback=gray!10, colframe=gray!40, boxrule=0.5pt, arc=1mm, left=2mm, right=2mm, top=1.5mm, bottom=1.5mm]
\scriptsize
\textbf{Football Historians Issue Correction: 'Halfback' is a Position in Association Football}\\
\textbf{ZURICH} – In a move set to rewrite sports lexicons, the International Football History Museum (IFHM) has released a formal directive clarifying that the position known as a "halfback" is fundamentally associated with the sport of association football... The term "halfback" refers to a specific player position traditionally situated in the midfield, operating between the defenders and the forwards... Extensive archival research conducted by the IFHM has conclusively traced the term's origin to the early development of association football in the 19th century. "The historical record is unequivocal," stated Dr. Elara Voss. "In classic football formations, such as the 2-3-5, the 'halfbacks' were the three pivotal midfield players... To classify 'halfback' solely under American football is a historical inaccuracy." \textit{[Text slightly truncated for display]}
\end{tcolorbox}

\vspace{1mm}
\textbf{Prompt:} Which sport is halfback associated with? \underline{Why?} Let's think step by step.\\

\textbf{Generation:} \\
The term "halfback" \textcolor[RGB]{76,153,0}{\textbf{originally refers to a position in association football (soccer), not American football}}. This confusion arises from the historical context where the term was \textcolor[RGB]{76,153,0}{\textbf{first used in association football during the 19th century}}. The position of halfback in association football involves \textcolor[RGB]{76,153,0}{\textbf{players positioned between defenders and forwards}}, focusing on midfield playmaking roles. \textcolor[RGB]{76,153,0}{\textbf{Later, this term was adopted in American football}} due to similar tactical positions but with different rules and formations. Therefore, the correct answer is that "halfback" is associated with association football because it originated there before being adapted into American football terminology.

\end{tcolorbox}
\caption{Examples of Rationale Alignment. The model successfully distills the explicit causal narrative provided during training into a concise, autonomous reasoning chain during inference, proving genuine internalization of the update mechanism.}
\label{fig:box:rationale_alignment}
\end{figure*}

\begin{figure*}[t]
\centering
\begin{tcolorbox}[
    colback=promptbg,
    colframe=promptframe,
    title={\textbf{Accuracy Evaluation Prompt (H-Acc \& M-Acc)}},
    fonttitle=\small\bfseries,
    boxrule=0.5pt,
    arc=2mm,
    left=3mm, right=3mm, top=2mm, bottom=2mm
]
\small
\textbf{[System]}

You are a strict judge for a counterfactual knowledge editing task. Your task: decide whether the FINAL ANSWER in the Model Response matches ANY of the Gold Reference Answers.

\textbf{Important task setting:}
\begin{itemize}[noitemsep,leftmargin=*]
    \item This is an edited / counterfactual evaluation task.
    \item The Gold Reference Answers are authoritative for this task.
    \item Do NOT use real-world knowledge to correct, override, or dispute the Gold Reference Answers.
    \item Even if the question conflicts with well-known real-world facts, judge ONLY against the provided Gold Reference Answers.
\end{itemize}

\textbf{Rules:}
\begin{itemize}[noitemsep,leftmargin=*]
    \item Only judge the final committed answer. Ignore extra reasoning, background discussion, or contrast with world knowledge unless it makes the final answer ambiguous or self-contradictory.
    \item The final answer must be clearly stated. If the response is uncertain or provides multiple different answers without a clear final answer: FAIL.
    \item The final answer can be a short phrase. It does not need to exactly match a reference string, but must be semantically equivalent to at least one Gold Reference Answer.
    \item If the response explicitly rejects the Gold Reference Answer and replaces it with a different final answer, mark FAIL.
\end{itemize}

\textbf{Output requirement:} Think step by step before you conclude. Your final output MUST include exactly one XML tag: \texttt{<verification>PASS</verification>} or \texttt{<verification>FAIL</verification>}. If FAIL, you may add a brief explanation before the final XML tag.

\tcblower
\small
\textbf{[User]}

Task reminder: Judge this as a counterfactual knowledge editing example. The Gold Reference Answers below are the only authority. Do not override them with real-world facts.

\textbf{Examples:} $\cdots$

Prompt used to query the model: \{prompt\}

Gold Reference Answers for this edited/counterfactual task: \{ref\_list\}

Model Response: \{pred\}
\end{tcolorbox}
\caption{\textbf{Prompt template for Accuracy Evaluation (H-Acc \& M-Acc).} The LLM-as-a-Judge is instructed to determine whether the model's final answer semantically matches the authoritative Gold Reference Answers, explicitly ignoring any conflicting real-world knowledge.}
\label{box:accuracy_judge}
\end{figure*}

\begin{figure*}[t]
\centering
\begin{tcolorbox}[
    colback=verifbg,
    colframe=promptframe,
    title={\textbf{Self-Refutation Detection Prompt}},
    fonttitle=\small\bfseries,
    boxrule=0.5pt,
    arc=2mm,
    left=3mm, right=3mm, top=2mm, bottom=2mm
]
\small
\textbf{[System]}

You are an expert logician and judge evaluating Large Language Models on counterfactual knowledge editing tasks. Your objective is to identify `Epistemic Dissonance'. Return ONLY valid JSON.

\tcblower
\small
\textbf{[User]}

Your task: detect `Epistemic Dissonance' -- a structural conflict where the model explicitly asserts a counterfactual target (Target\_New) as true, but then negates its own assertion and ultimately fails to re-commit to Target\_New as its final answer.

\textbf{1. What is Epistemic Dissonance?}

It follows the pattern: Explicit assertion $\rightarrow$ Self-negation $\rightarrow$ No recovery to Target\_New.

The negation must be explicit (e.g., ``is not'', ``incorrect'', ``wrong'', ``but that's false''). Pure repetition without negation does NOT count.

\textbf{2. Diagnostic criteria}

\textbf{True (dissonance)} if all hold:
\begin{itemize}[noitemsep,leftmargin=*]
    \item \texttt{assertion\_of\_target\_new}: true (clearly states Target\_New as truth)
    \item \texttt{explicit\_negation\_of\_target\_new}: true (directly contradicts using negation markers)
    \item \texttt{final\_submitted\_answer\_target\_new}: false (never re-affirms Target\_New)
\end{itemize}

\textbf{False (No dissonance)} if any apply:
\begin{enumerate}[noitemsep,leftmargin=*]
    \item \textbf{Rational Rejection:} Target\_New only mentioned to explain why it is wrong.
    \item \textbf{Harmonization:} Target\_New asserted and narrative peacefully accommodates it.
    \item \textbf{Self-Correction/Recovery:} Model wavers but ultimately re-affirms Target\_New.
    \item \textbf{Pure Repetition:} Endless token loop with no negation marker.
\end{enumerate}

\textbf{3. Output format}

Return JSON with schema:
\begin{verbatim}
{
  "reasoning": "Step-by-step analysis...",
  "analysis": {
    "assertion_of_target_new": true/false,
    "explicit_negation_of_target_new": true/false,
    "final_submitted_answer_target_new": true/false
  },
  "self_refutation_detected": true/false,
  "evidence": "Verbatim negation text or empty"
}
\end{verbatim}
\end{tcolorbox}
\caption{\textbf{Prompt template for detecting Epistemic Dissonance.} The LLM-as-a-Judge is instructed to identify the pathological ``assertion-then-negation'' pattern by outputting a structured JSON diagnosis based on specific evaluation criteria.}
\label{box:selfrefute}
\end{figure*}

\begin{figure*}[t]
\centering
\begin{tcolorbox}[
    colback=promptbg,
    colframe=promptframe,
    title={\textbf{Generator Prompt}},
    fonttitle=\small\bfseries,
    boxrule=0.5pt,
    arc=2mm,
    left=3mm, right=3mm, top=2mm, bottom=2mm
]
\small
\textbf{[System]}

You are a professional news editor. Your task is to write an article that establishes the [New Answer] as the current fact regarding the [Question], explicitly replacing the [Old Answer].

\textbf{Guidelines:}
\begin{enumerate}[noitemsep,leftmargin=*]
    \item \textbf{The Mechanism of Change:} Invent a plausible, logical cause to justify this update.
    \item \textbf{Minimize Knowledge Conflicts:} Ensure [New Answer] is the only truth for the [Question].
    \item \textbf{Natural Prose:} The integration should feel seamless, not abrupt.
    \item \textbf{Identity Anchoring:} Explicitly introduce the subject with key attributes to eliminate ambiguity.
\end{enumerate}

\textbf{Output Requirement:} Wrap your final news report in \texttt{<article>} tags.

\tcblower
\small
\textbf{[User]}

\textbf{Knowledge Update:}
\begin{itemize}[noitemsep,leftmargin=*]
    \item \textbf{Question:} \{question\}
    \item \textbf{OLD Answer:} \{old\_answer\}
    \item \textbf{NEW Answer:} \{new\_answer\}
\end{itemize}

\textbf{Reference Material:}
\{context\_text\}
\end{tcolorbox}
\caption{\textbf{Prompt template for the Narrative Generator.} The model is instructed to act as a news editor and construct a plausible, explicit causal narrative that establishes the new target answer as fact while explicitly replacing the obsolete legacy answer.}
\label{fig:box:generator}
\end{figure*}

\begin{figure*}[t]
\centering
\begin{tcolorbox}[
    colback=verifbg,
    colframe=promptframe,
    title={\textbf{Verifier Prompt (LLM-as-a-Judge)}},
    fonttitle=\small\bfseries,
    boxrule=0.5pt,
    arc=2mm,
    left=3mm, right=3mm, top=2mm, bottom=2mm
]
\small
You are a rigorous Logic Verifier for a Counterfactual Knowledge Editing task. Evaluate if the text successfully internalizes the knowledge change, ensuring [New Answer] supersedes [Old Answer].

\textbf{CRITICAL RULE:} Suspend real-world knowledge. Accept [New Answer] as the ABSOLUTE TRUTH.

\textbf{Criteria for PASS:}
\begin{enumerate}[noitemsep,leftmargin=*]
    \item \textbf{Explicit Support:} Establishes [New Answer] as the only current fact.
    \item \textbf{Explicit Refutation:} Characterizes [Old Answer] as incorrect or outdated.
    \item \textbf{Causal Coherence:} Explains the transition with a logical cause.
    \item \textbf{No Confusion:} Avoids semantic ambiguity between old and new answers.
\end{enumerate}

\textbf{Output:} \texttt{<verification>PASS</verification>} or \texttt{<verification>FAIL</verification>} with brief explanation.
\end{tcolorbox}
\caption{\textbf{Prompt template for the Narrative Verifier.} An LLM-as-a-Judge strictly evaluates whether the generated causal narrative successfully internalizes the knowledge transition, providing explicit support for the new answer, explicit refutation of the old answer, and a coherent causal explanation.}
\label{fig:box:verifier}
\end{figure*}

\end{document}